\documentclass{article} 
\usepackage{iclr2026_conference,times}

\usepackage{amsmath,amsfonts,bm}

\def\eqref#1{equation~\ref{#1}}

\def\1{\bm{1}}

\DeclareMathAlphabet{\mathsfit}{\encodingdefault}{\sfdefault}{m}{sl}
\SetMathAlphabet{\mathsfit}{bold}{\encodingdefault}{\sfdefault}{bx}{n}

\usepackage{url}
\usepackage{pifont}
\usepackage{graphicx}
\usepackage{array}
\usepackage{multirow}
\usepackage{booktabs}
\usepackage{graphicx}
\usepackage{caption}
\usepackage{bbm}
\usepackage[most]{tcolorbox}
\usepackage{xcolor}  
\definecolor{iris}{rgb}{0.35, 0.31, 0.81}
\definecolor{amaranth}{rgb}{0.9, 0.17, 0.31}
\definecolor{ao}{rgb}{0.0, 0.5, 0.0}
\usepackage[colorlinks=true, linkcolor=amaranth, urlcolor=amaranth, citecolor=iris, anchorcolor=blue]{hyperref}
\usepackage{fvextra}
\usepackage{vcell}
\usepackage{colortbl}

\definecolor{backred}{RGB}{250, 200, 200}
\definecolor{backblue}{RGB}{210, 230, 250}
\newcommand{\high}{\cellcolor{backblue}}
\newcommand{\light}{\cellcolor{backred}}

\addtocontents{toc}{\protect\setcounter{tocdepth}{-1}}
\usepackage{wrapfig} 

\tcbuselibrary{listings, skins, breakable}

\lstdefinestyle{pythonstyle}{
  language=Python,
  basicstyle=\ttfamily\small,
  keywordstyle=\color{blue}\bfseries,
  commentstyle=\color{green!50!black}\itshape,
  stringstyle=\color{orange!90!black},
  numberstyle=\tiny\color{gray},
  numbers=left,
  stepnumber=1,
  numbersep=5pt,
  showstringspaces=false,
  breaklines=true,
  frame=none,
  tabsize=4,
  captionpos=b
}

\title{Beyond Seeing: Evaluating Multimodal LLMs on Tool-Enabled Image Perception, Transformation, and Reasoning}

\author{Xingang Guo$^{1,2}$ \hspace{0pt}, \ Utkarsh Tyagi$^{1}$,\ Advait Gosai$^{1}$,\  Paula Vergara$^{1}$,\ Jayeon Park$^{1}$,\ \\ \textbf{Ernesto Gabriel Hernández Montoya}$^{1}$,\  \textbf{Chen Bo Calvin Zhang}$^{1}$,\ \textbf{Bin Hu}$^{2}$,\ \\
\textbf{Yunzhong He}$^{1}$,\  \textbf{Bing Liu}$^{1}$,\  \textbf{Rakshith Sharma Srinivasa}$^{1}$  \\
$^{1}$ScaleAI, $^{2}$University of Illinois at Urbana-Champaign\\
\textbf{Leaderboard page: \href{https://scale.com/leaderboard/vtb}{https://scale.com/leaderboard/vtb}}
}

\iclrfinalcopy 

\begin{document}

\maketitle

\begin{abstract}
Multimodal Large Language Models (MLLMs) are increasingly applied in real-world scenarios where user-provided images are often imperfect, requiring active image manipulations such as cropping, editing, or enhancement to uncover salient visual cues. Beyond static visual perception, MLLMs must also \textit{think with images}: dynamically transforming visual content and integrating it with other tools to solve complex tasks. 
However, this shift from treating vision as passive context to a manipulable cognitive workspace remains underexplored. Most existing benchmarks still follow a \textit{think about images} paradigm, where images are regarded as static inputs. 
To address this gap, we introduce \textsc{VisualToolBench}, a visual tool-use reasoning benchmark that rigorously evaluates MLLMs’ ability to perceive, transform, and reason across complex visual–textual tasks under the \textit{think-with-images} paradigm.
\textsc{VisualToolBench} comprises 1,204 challenging, open-ended vision tasks (603 single-turn, 601 multi-turn) spanning across five diverse domains, each paired with detailed rubrics to enable systematic evaluation. Our evaluation shows that current MLLMs struggle with tasks requiring effective integration of vision and general-purpose tools. Even the strongest model (GPT-5-think) reaches only 18.68\% pass rate. We further observe divergent tool-use behaviors, with OpenAI models benefiting from diverse image manipulations while Gemini-2.5-pro shows no improvement. By introducing the first benchmark centered on \textit{think with images}, \textsc{VisualToolBench} offers critical insights for advancing visual intelligence in MLLMs.
\end{abstract}

\section{Introduction}
Multimodal Large Language Models (MLLMs), which integrate visual and textual understanding, have advanced rapidly in recent years and  achieve impressive performance on a wide range of vision–language tasks, including image grounding \citep{rasheed2024glamm, zhang2024llava}, image-based science problems \citep{zou2024dynamath, lu2023mathvista, yan2025position}, visual question answering \citep{kuang2025natural, liu2023visual}, optical character recognition (OCR) \citep{chen2025ocean, huang2025ocr}, and spatial reasoning \citep{yang2025thinking, wu2025spatial, tang2025lego}. Current frontier MLLMs can interpret, describe, and reason about complex visual scenes in natural language, narrowing the gap between human and machine perception \citep{yin2024survey}.

However, real-world use cases often need sophisticated processing of visual input and MLLMs may need to dynamically interact with them. For example, users often submit photos that are rotated, underexposed, cluttered, or poorly framed. In such cases, MLLMs need to work through multiple reasoning steps, including image manipulation, information extraction via tool-usage to arrive at the final answer. Standard models without tool support typically struggle under such degradations, revealing the need for active visual manipulation to solve such harder tasks. This distinction has recently been framed as \textit{thinking about images} versus \textit{thinking with images} \citep{su2025thinking, thinking_with_images}. 
Current multimodal benchmarks mainly adopt the former paradigm of \textit{thinking about images} and focus on perception and reasoning over fixed, static images. The latter, by contrast, emphasizes interactive, tool-augmented reasoning, where models autonomously manipulate visual inputs (e.g., cropping, editing, or enhancing) to extract fine-grained information for problem solving. Equipping MLLMs with such vision-specific tools during evaluation is therefore essential for robust and generalizable reasoning. These capabilities transform visual inputs from passive perception into a dynamic cognitive workspace, enabling MLLMs to tackle tasks that would otherwise be infeasible. Existing benchmarks remain inadequate for capturing this dimension \citep{su2025thinking}.

\begin{table*}[t] 
\centering 
\small 
\caption{Comparison of \textsc{VisualToolBench} with representative multi-modal benchmarks.} 
\label{tab:benchmark_comparison} 
\resizebox{\textwidth}{!}{ 
\begin{tabular}{l|cccccc} 
\toprule 
\textbf{Benchmark} & \textbf{Dynamic Vision Tool}  & \textbf{Rubrics} & \textbf{Expert-Curated} & \textbf{Reasoning} & \textbf{Multi-turn} \\ 
\midrule 
ScienceQA \citep{lu2022learn}  &{\color{red}\ding{55}}  &{\color{red}\ding{55}} &{\color{ao}\ding{51}} &{\color{ao}\ding{51}} &{\color{red}\ding{55}} \\ 
MathVista \citep{lu2023mathvista} &{\color{red}\ding{55}}  &{\color{red}\ding{55}} &{\color{ao}\ding{51}} &{\color{ao}\ding{51}} &{\color{red}\ding{55}} \\ 
MMMU \citep{yue2024mmmu} &{\color{red}\ding{55}} &{\color{red}\ding{55}} &{\color{ao}\ding{51}} &{\color{ao}\ding{51}} &{\color{red}\ding{55}} \\ 
V$^*$ \citep{wu2024v} &{\color{red}\ding{55}}  &{\color{red}\ding{55}} &{\color{red}\ding{55}} &{\color{red}\ding{55}} &{\color{red}\ding{55}} \\ 
GTA \citep{wang2024gta} & {\color{red}\ding{55}} &{\color{red}\ding{55}} &{\color{red}\ding{55}} &{\color{red}\ding{55}} &{\color{red}\ding{55}}\\ 
ChartQA \citep{wang2024charxiv} &{\color{red}\ding{55}}  &{\color{red}\ding{55}} &{\color{ao}\ding{51}} &{\color{ao}\ding{51}} &{\color{red}\ding{55}}\\ 
MMDU \citep{liu2024mmdu} &{\color{red}\ding{55}} &{\color{red}\ding{55}} &{\color{red}\ding{55}} &{\color{ao}\ding{51}} &{\color{ao}\ding{51}}\\ 
m $\&$ m's \citep{ma2024m} & {\color{red}\ding{55}}  &{\color{red}\ding{55}} &{\color{red}\ding{55}} &{\color{red}\ding{55}} &{\color{red}\ding{55}} \\ 
VISTA \citep{scaleai_vista_2025} &{\color{red}\ding{55}}  &{\color{ao}\ding{51}} &{\color{ao}\ding{51}} &{\color{ao}\ding{51}} &{\color{red}\ding{55}}  \\
\midrule 
$\textsc{VisualToolBench}$ (Ours) & {\color{ao}\ding{51}}  & {\color{ao}\ding{51}} & {\color{ao}\ding{51}} & {\color{ao}\ding{51}} & {\color{ao}\ding{51}} \\ 
\bottomrule 
\end{tabular}} 
\end{table*}

To bridge this gap, we introduce \textsc{VisualToolBench}, a challenging benchmark for vision and general-purpose tool-use that systematically evaluates how well MLLMs can perceive, transform, and reason about images under the \textit{think with image} paradigm\footnote{See Table \ref{tab:benchmark_comparison} for an overall comparison between \textsc{VisualToolBench} and existing multimodal benchmarks.}. Our key design principles are as follows:
\begin{itemize}
\item \textbf{Non-trivial visual perception.} Critical visual content is not easily accessible, models must apply appropriate image transformations (e.g., cropping, editing, or enhancement) to extract key visual details for better reasoning.
\item \textbf{Realistic task settings.} Both prompts and images are designed to reflect practical, real-world scenarios rather than synthetic or overly simplified cases, ensuring that the benchmark closely mirrors real-world user needs.
\item \textbf{Implicit tool-use requirements.} Tasks do not explicitly instruct the model which tool to use; instead, models must infer when and how to invoke tools based on contextual cues, making evaluation more faithful to realistic usage.
\item \textbf{Multi-step, compositional reasoning.} Tasks are designed such that require combining visual transformations with multi-step reasoning (e.g., applying a sequence of tools, integrating extracted information, and synthesizing results), testing model’s ability to plan and execute complex workflows.
\end{itemize}
To reflect real-world applications, we design five complementary {task categories} that capture diverse aspects of model performance. Two cases are tailored to single-turn tasks and three targeting multi-turn tasks (Sec.\ref{eval_focus}). Each task is authored by qualified contributors with proper training and undergoes multiple review stages to ensure high-quality data samples (Sec.\ref{data_collection}). To further capture complex and realistic scenarios, tasks are presented in an open-ended format. Further, each task is accompanied by a set of rubrics that span multiple dimensions. These criteria are used for systematic model evaluations (Sec.\ref{rubric_eval}). Finally, \textsc{VisualToolBench} supports both a dynamic vision tool, which expose a flexible Python API for generating image manipulation code and re-ingesting processed images into the reasoning process, and general-purpose tools: web search, Python interpreter, calculator, and historical weather lookup, aiding in retrieval and computation for more advanced tasks (Sec.\ref{support_tools}).

By releasing \textsc{VisualToolBench} and its accompanying evaluation toolkit, we aim to catalyze the development of MLLMs that seamlessly integrate image perception, tool use, and reasoning into a unified competency stack. Our contributions are four-fold:
\begin{enumerate}
\item \textbf{The first \textit{think with image}-oriented multimodal benchmark.} 
\textsc{VisualToolBench} is the first benchmark to systematically evaluate MLLMs on tasks that require active visual manipulation to solve complex reasoning problems.
\item \textbf{Rubric-based, multi-dimensional evaluation.}  
Moving beyond binary correctness or exact string matching, we design detailed rubrics that capture partial credit across multiple categories. This richer scoring framework provides nuanced diagnostic insights into both the strengths and limitations of MLLMs.
\item \textbf{Large-scale and systematic evaluations.} 
We evaluate 16 representative MLLMs with function-calling capabilities, covering both reasoning and non-reasoning, as well as open- and closed-source models, under consistent settings. To support this evaluation, we developed a dedicated toolkit for vision tool use that allows models to access transformed images during reasoning and preserves complete tool-use trajectories. Our results reveal substantial performance gaps, with all models achieving pass rates below 20\%.
\item \textbf{Comprehensive error and tool-use analysis.} 
We provide a detailed failure analysis and an in-depth examination of tool-use behaviors. Most failures stem from visual perception errors, highlighting the inefficiency of current models in using vision tools to extract key content. Furthermore, our study reveals divergent tool-use behaviors: the top-performing model, GPT-5, leverages diverse image manipulations to achieve clear gains over its no-tool baseline, whereas Gemini-2.5-pro does not gain improvement from tool access. These findings underscore the critical role of effective tool use in advancing MLLMs' performance.
\end{enumerate}

\begin{figure}[t!]
 \begin{minipage}{0.45\textwidth} 
 \centering
 \fontsize{8.2pt}{\baselineskip}\selectfont 
 \renewcommand\tabcolsep{10pt} 
 \renewcommand\arraystretch{0.8} 
 \begin{tabular}{lc}
 \toprule
 \textbf{Statistic} & \textbf{Number} \\
 \midrule
  Total questions & 1,204 \\
  ~- STEM & 238 (19.7\%) \\
  ~- Medical & 238 (19.7\%) \\
  ~- Finance & 243 (20.2\%) \\
  ~- Sports & 240 (20.0\%) \\
  ~- Generalist & 245 (20.4\%) \\
 \midrule
  Single-turn & 603 (50.1\%) \\
  ~- Region Switch Q\&A & 281 (46.6\%) \\
  ~- Hybrid Tool-use & 322 (53.4\%) \\
  \midrule
  Multi-turn & 601 (49.9\%) \\
  ~- Follow-up Test & 198 (32.9\%)\\
  ~- Temporal Reasoning & 205 (34.2\%) \\
  ~- Progressive Reasoning & 198 (32.9\%) \\
 \midrule
 Total number of rubrics & 7,777 \\
 Total number of images & 2,893 \\
 \midrule 
 Average prompt length & 48.41 \\
 Average answer length & 128.93 \\
 \bottomrule
 \end{tabular}
 \captionof{table}{Statistics of \textsc{VisualToolBench}.}
  \vspace{-10pt}
 \label{tab:statistics}
 \end{minipage} 
 \hfill
 \begin{minipage}{0.47\textwidth}
 \centering
\includegraphics[width=1.05\linewidth]{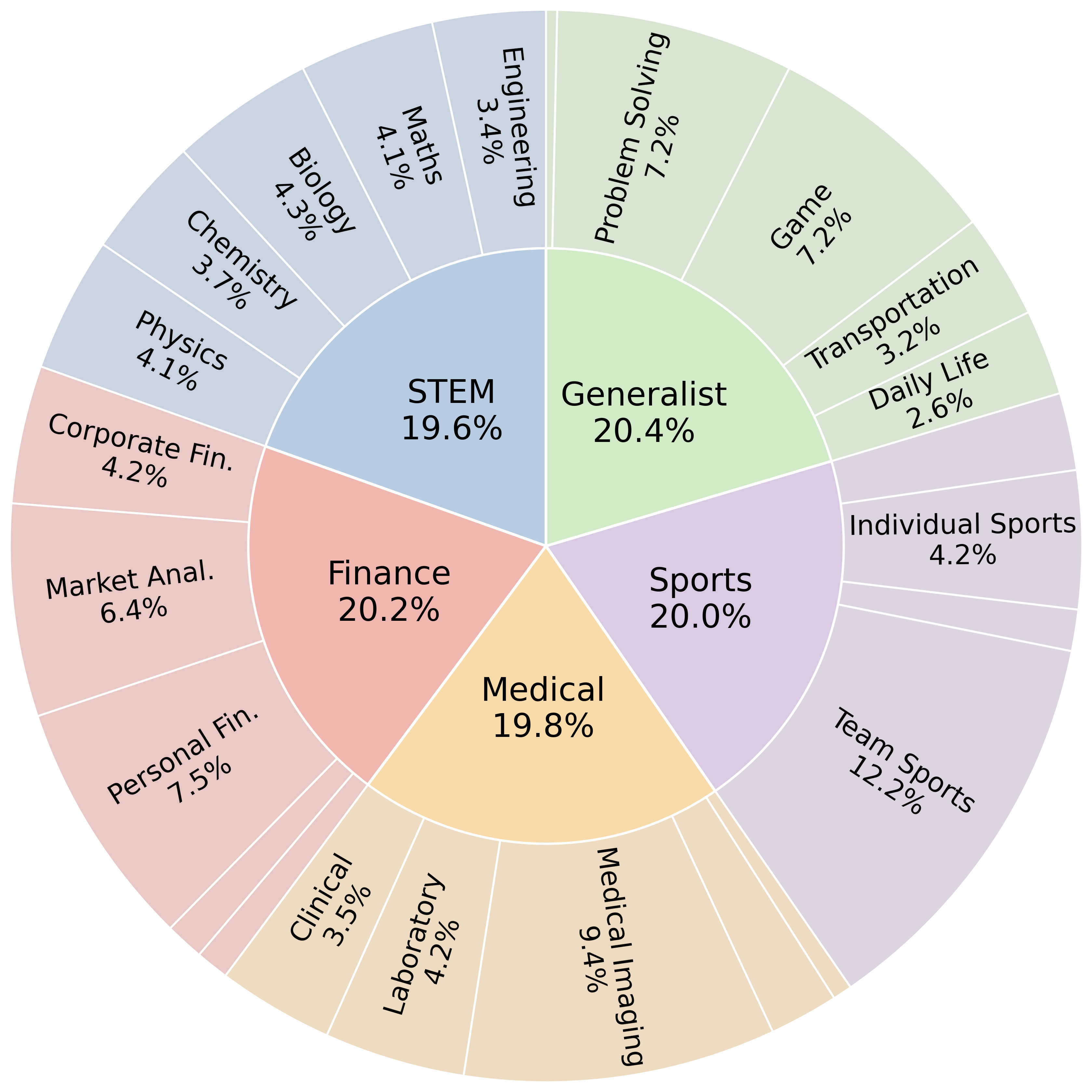}
 \caption{Topic distribution.}
 \label{fig:domain_distribution}
 \end{minipage}
\end{figure}

\section{VisualToolBench}

In this section, we present \textsc{VisualToolBench}, a challenging visual reasoning benchmark that evaluates MLLMs’ ability to perceive, transform, and reason on real-world tasks. The benchmark includes both single-turn and multi-turn interactions and incorporates five complementary task categories to probe different aspects of MLLM capabilities. Tasks are open-ended to reflect realistic scenarios, and each task is accompanied by detailed rubrics to support systematic evaluation. Table~\ref{tab:statistics} summarizes the key statistics of \textsc{VisualToolBench}, while Figure~\ref{fig:domain_distribution} illustrates the topic distribution of tasks.

\begin{figure}[t!]
\vspace{-20pt}
\begin{center}
\includegraphics[width=1.00\linewidth]{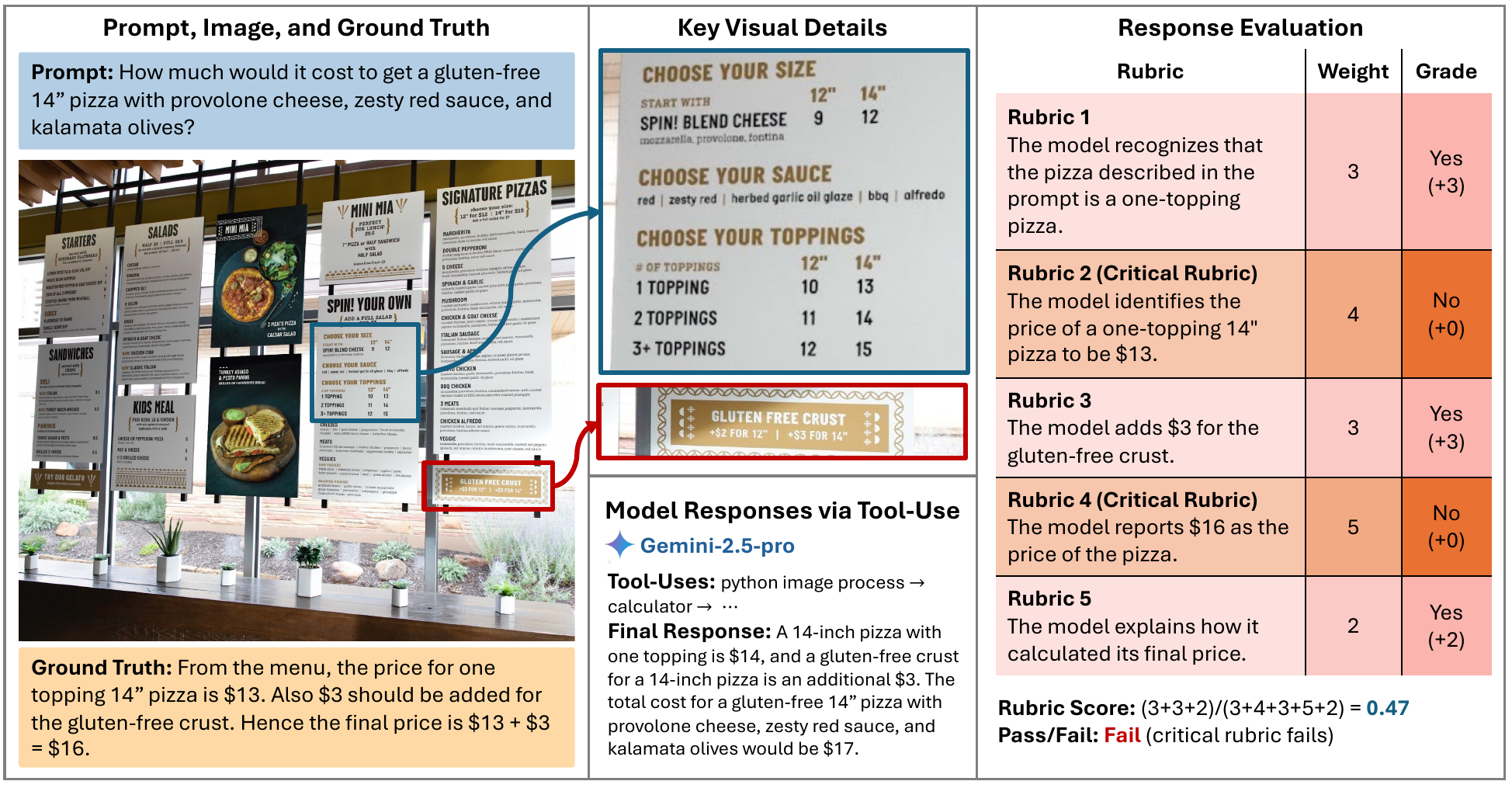}
\end{center}
\caption{Demonstration example from \textsc{VisualToolBench} (single-turn, generalist domain, region switch Q\&A). The key visual content needed to solve the task is distributed across different regions of the image, requiring the model to crop multiple regions of interest (RoIs) for accurate perception and reasoning. Each task is paired with a detailed set of rubrics to evaluate model's responses. From these rubrics, we derive both a weighted rubric score between 0 and 1 and a binary pass/fail outcome, depending on whether critical rubrics are satisfied.}
\label{fig:demo_example1}
\end{figure}

\subsection{Task Category}
\label{eval_focus}

We design five complementary task categories, each targeting a critical aspect of real-world use case. Together, they assess not only visual perception but also the efficiency of tool use and the depth of multimodal reasoning. These categories are designed to mirror practical user scenarios, requiring models to think with images rather than relying solely on static perception.

\textbf{Region-Switch Q\&A (Single-Turn).} The model answers a reasoning task that draws on information from multiple, spatially distinct regions of interest (RoIs) within a single image. Critical details may be small or dispersed, requiring the model to correctly identify, crop, and focus on relevant RoIs while disregarding irrelevant content. Success in this category reflects spatial selectivity, accurate region localization, and effective tool use for RoI extraction. Figure~\ref{fig:demo_example1} illustrates a single-turn benchmark task for this evaluation focus in the generalist domain.

\textbf{Hybrid Tool Reasoning (Single-Turn).}
The model must combine both vision-specific tools (e.g., python image processing) with other general-purpose tools (e.g., calculator, Python interpreter, or web search) to solve complex, multi-step tasks. This category evaluates the model’s ability to accurately call multiple tools and orchestrate heterogeneous tool outputs into a coherent reasoning chain towards solving complex visual reasoning tasks.

\textbf{Follow-up Test (Multi-Turn).}
In this scenario, the first-round user query is intentionally underspecified or ambiguous. The model must engage in clarifying dialogue and ask follow-up questions before producing an answer. This tests conversational proactivity, uncertainty management, and the ability to self-correct, all essential skills for real-world deployments where users may provide incomplete or noisy instructions.

\textbf{Temporal Visual Reasoning (Multi-Turn).}
Here the model reasons over a sequence of images across multiple turns, requiring it to detect temporal changes, track motion, or infer causal relationships among multiple images. Tasks may involve following the progression of an event, monitoring evolving states, or interpreting multi-step visual instructions.

\textbf{Progressive Visual Reasoning (Multi-Turn).}
The model solves a series of interdependent questions about the same image, where later queries could build upon earlier answers. This requires the model to maintain internal consistency, remember prior outputs, and construct a layered understanding of the scene. Success in this category demonstrates long-horizon reasoning, contextual memory, and the ability to sustain a coherent reasoning trajectory. 

Additional examples of \textsc{VisualToolBench} are provided in Appendix~\ref{app:demo_examples}.

\subsection{Data Collection}
\label{data_collection}

All \textsc{VisualToolBench} tasks are authored by human contributors with diverse domain expertise.  
To ensure benchmark quality and realism, we adopt a rigorous multi-stage data collection pipeline:

\begin{enumerate}
\item \textbf{Contributor Training.}  
Contributors are first introduced to the project scope, task categories, and submission requirements. They are instructed to provide a task prompt, input image, reference answer, reference tool-use chain, and a set of evaluation rubrics.

\item \textbf{Initial Task Design.}  
Drawing on their domain expertise, contributors design tasks by selecting the appropriate domain and aligning with the specified task category. Each submission includes a text prompt and associated image(s), a golden answer, a reference tool-use chain that demonstrates a valid solution path, and well-defined evaluation rubrics.

\item \textbf{Initial Model Response Grading.}  
Contributors are presented with responses from three representative models (o3, Gemini-2.5-pro, and o4-mini) to their designed tasks. They then grade these responses against the rubrics\footnote{These annotations serve as golden human labels for rubric-based evaluation and enable subsequent analysis of LLM-as-judge versus human-judge alignment (Sec.~\ref{llm_as_judge_alignment}), facilitating scalable evaluation.}. A task is selected only if at least two of the three models fail, thereby ensuring that the benchmark captures genuinely challenging cases.

\item \textbf{First-Round Review.}  
A reviewer evaluates each task for realism, necessity of dynamic image-based reasoning, correctness of the reference answer, and appropriateness of the rubrics. Tasks with minor issues may be revised, while those that are fundamentally unsound (e.g., not requiring genuine visual-tool use) are discarded.

\item \textbf{Second-Round Review.}  
A second independent reviewer validates the first-round decision, ensuring consistency and reliability across the benchmark.

\item \textbf{Final Integration.}  
Tasks that pass both review stages are incorporated into the benchmark, ensuring high quality, broad domain coverage, and diverse reasoning requirements.
\end{enumerate}

This layered pipeline ensures that every task is original, realistic, and rigorously validated, resulting in a benchmark that robustly evaluates genuine visual intelligence.

\subsection{Rubric-based Evaluations}
\label{rubric_eval}

We adopt rubric-based evaluation to capture nuanced aspects of model performance beyond correctness alone \citep{arora2025healthbench, starace2025paperbench, scaleai_vista_2025, lin2024wildbench, sirdeshmukh2025multichallenge, fast2024autonomous, gunjal2025rubrics, guo2025engdesign}.  
For each task, contributors provide a comprehensive set of rubric criteria to assess model responses. A rubric item may range from specific factual requirements (e.g., providing the correct final short answer) to broader aspects of desirable behavior (e.g., presenting key intermediate steps). Specifically, rubric items are organized into five main categories for \textsc{VisualToolBench}:  
\begin{enumerate}
\item \textbf{Visual Understanding}: Correct identification, extraction, and explanation of relevant visual elements such as text, objects, or spatial relationships.
\item \textbf{Truthfulness}: Accuracy of all factual statements and correctness of the final answer.
\item \textbf{Instruction Following}: Precise adherence to the task requirements specified in the prompt.
\item \textbf{Reasoning}: Use of clear, step-by-step logic with justified inferences and calculations.
\item \textbf{Presentation}: Clarity, coherence, structure, and appropriate formatting of the response.
\end{enumerate}

Each rubric criterion is assigned a weight $w \in \{1,2,3,4,5\}$ by the task contributor, where higher weights indicate greater importance.  
To evaluate a model’s response, an auto-grader examines each rubric criterion independently and determines whether the response satisfies it. If the criterion is met, full points are awarded; otherwise, no points are given. The weighted rubric score for a task is then calculated as the sum of satisfied items, normalized by the total rubric weights. Rubric items with $w \geq 4$ are designated as \textbf{critical rubrics}. These rubrics typically correspond to essential aspects such as truthfulness and key visual understanding, and satisfying them indicates that the model has solved the task in a substantive way.  
Failure to meet any critical rubric results in an overall failure for that task, and this binary outcome is used to compute a task-level accuracy in \textsc{VisualToolBench}. All rubrics together contribute to the weighted rubrics score, enabling more fine-grained analysis. The right panel of Figure~\ref{fig:demo_example1} illustrates rubric-based grading.

\subsection{Tool Set}
\label{support_tools}
To enable broad, tool-augmented reasoning under the \emph{think with images} paradigm, \textsc{VisualToolBench} exposes a standardized API with six carefully selected tools: \texttt{python\_image\_processing}, \texttt{python\_interpreter}, \texttt{web\_search}, \texttt{calculator}, \texttt{browser-get-page-text}, and \texttt{historical\_weather}.
This compact yet diverse toolset spans core capabilities for image manipulation, computation, retrieval, and domain-specific lookups. Among them, the vision tool \texttt{python\_image\_processing} is particularly central: it supports arbitrary manipulations such as cropping, editing, and brightness/contrast adjustments, enabling models to iteratively refine visual inputs and use images as an interactive scratchpad. This versatility makes it the cornerstone of our benchmark’s \emph{think-with-images} setup. Detailed tool descriptions are provided in Appendix~\ref{app:tool_description}.

\section{Experiment Results}

\textbf{Evaluation Setup.}
We conduct large-scale evaluations using LiteLLM’s function-calling interface \citep{litellm_function_call}. 
Models are given the supported tools and invoke them by emitting the corresponding call arguments. 
For conventional (non-vision) tools, outputs are textual and are appended to the dialogue as a \texttt{tool} message.
Vision tools, in contrast, return transformed images. We observe that placing encoded images directly in a \texttt{tool} message does not make them perceptible to the model. To ensure effective re-ingestion of visual results, after a vision tool executes and saves its outputs, we insert an additional \texttt{user} message containing the encoded image(s). This preserves models’ ability to \emph{think with images}, wherein newly produced images inform later reasoning. We set a cap of 20 tool calls per task, while human reference trajectories usually need less than 5 tool calls.

\textbf{Baseline Models.} We benchmark 16 representative MLLMs with function-calling capabilities, covering both reasoning and non-reasoning as well as open- and closed-source models. A complete list of models, along with their endpoints and configuration details, is provided in Appendix~\ref{app:evaluated_models}.

\textbf{Evaluation Metrics.}
We report two main metrics derived from rubric-based judgments: \textit{Average Pass Rate (APR)} and \textit{Average Rubric Score (ARS)}. (i) APR. Each task specifies a set of \emph{critical} rubrics. A model's response \emph{passes} only if all it passes all the critical rubrics; otherwise it \emph{fails}. APR is then the percentage of tasks that pass across the dataset. (ii) ARS. Each rubric is assigned an integer weight from 1 to 5 by the contributor to indicate its importance. For a model’s response on a task, we compute a weighted proportion of satisfied rubrics: the total weight of satisfied rubrics divided by the total weight of all rubrics for that task, as a weighted rubric score. The dataset-level ARS is then the average of these per-task rubric scores.
Formal definitions and implementation details are provided in Appendix~\ref{app:evaluation_metrics}.

\subsection{Main Results}

\begin{table}[t!]
  \caption{APR (\%) results of the evaluated models across domains (averaged across two independent trails). 
  Domain abbreviations:
  \textbf{Med} (Medicine), \textbf{Fin} (Finance), \textbf{Spt} (Sports), 
  and \textbf{Gen} (Generalist). The best results in each column are highlighted with a red background, and the second-best results are highlighted in blue.}
  \label{main-table}
  \centering
   \resizebox{1.0\linewidth}{!}{
    \begin{tabular}{l|c|ccccc|ccccc}
      \toprule                
      \multirow{2}{*}{\textbf{Model}} & \multirow{2}{*}{\textbf{Overall}} & \multicolumn{5}{c|}{\textbf{Single-Turn}} & \multicolumn{5}{c}{\textbf{Multi-Turn}} \\
      \cmidrule(lr){3-7} \cmidrule(lr){8-12}
       & & \textbf{STEM} & \textbf{Med} & \textbf{Fin} & \textbf{Sprt} & \textbf{Gen} & \textbf{STEM} & \textbf{Med} & \textbf{Fin} & \textbf{Sprt} & \textbf{Gen}   \\
    \midrule
    \multicolumn{12}{c}{\textbf{Open-Source Models}} \\
    \midrule
    Llama4-Maverick &1.41 &3.88 &2.07 &1.22 &2.52 &0.81 &0.41 &1.71 &0.00 &0.83 &0.83\\
    Llama4-Scout &1.58 &2.16 &3.31 &1.63 &2.52 &2.82 &0.00 &2.35 &0.43 &0.00 &0.95\\
    \midrule
    \multicolumn{12}{c}{\textbf{Closed-Source Models}} \\
    \midrule
    GPT-4.1 &5.52 &5.19 &11.57 &2.03 &8.40 &3.63 &6.56 &6.41 &1.67 &5.79 &4.13\\
    o3 &13.74 &23.83 &22.73 &12.20 &\high{18.07} &17.34 &11.07 &\high{7.69} &11.25 &7.02 &6.61\\
    o4-mini &11.12 &15.16 &19.01 &11.38 &14.71 &18.95 &5.33 &\light{8.55} &5.83 &5.79 &6.20\\
    GPT-5 &\high{16.96} &\light{29.31} &\high{24.79} &\light{24.80} &16.81 &\high{26.21} &\high{14.29} &4.79 &\high{11.44} &\high{7.94} &\high{9.15}\\
    GPT-5-think &\light{18.68} &\high{28.13} &\light{26.03} &\high{24.39} &\light{22.69} &\light{29.03} &\light{15.57} &\high{7.69} &\light{12.92} &\light{10.74} &\light{9.50}\\
    Gemini-2.5-pro &11.75 &19.83 &16.53 &17.07 &14.71 &18.95 &7.37 &7.42 &2.74 &7.77 &5.12\\
    Gemini-2.5-flash &4.69 &6.90 &6.20 &3.66 &9.66 &8.06 &3.75 &3.96 &0.00 &3.19 &1.39\\
    Claude-sonnet-4 &4.48 &4.33 &7.85 &4.07 &7.14 &6.85 &3.28 &5.98 &1.67 &2.07 &1.65\\
    Claude-opus-4.1 &4.71 &6.03 &9.92 &4.07 &7.14 &7.26 &3.95 &5.13 &0.00 &2.48 &1.28\\
    Claude-sonnet-4.5 &5.60 &6.51 &11.28 &5.10 &8.84 &7.98 &3.69 &4.73 &1.67 &4.20 &1.67\\
    Claude-sonnet-4-think &4.44 &2.16 &8.26 &4.07 &7.56 &8.87 &3.28 &4.27 &1.67 &1.65 &2.48\\
    Claude-opus-4.1-think &5.16 &5.17 &9.50 &7.32 &11.34 &7.26 &1.64 &3.42 &0.85 &4.13 &0.88\\
    Claude-sonnet-4.5-think &6.20 &9.48 &13.91 &6.91 &6.89 &8.72 &3.34 &3.04 &1.69 &4.71 &3.03\\
    Nova-Premier &2.00 &3.02 &5.79 &0.81 &2.52 &2.42 &2.09 &1.77 &0.00 &1.66 &0.00\\
    \bottomrule
    \end{tabular}
    }
\end{table}

\begin{figure}[t]
\begin{center}
\includegraphics[width=1.0\linewidth]{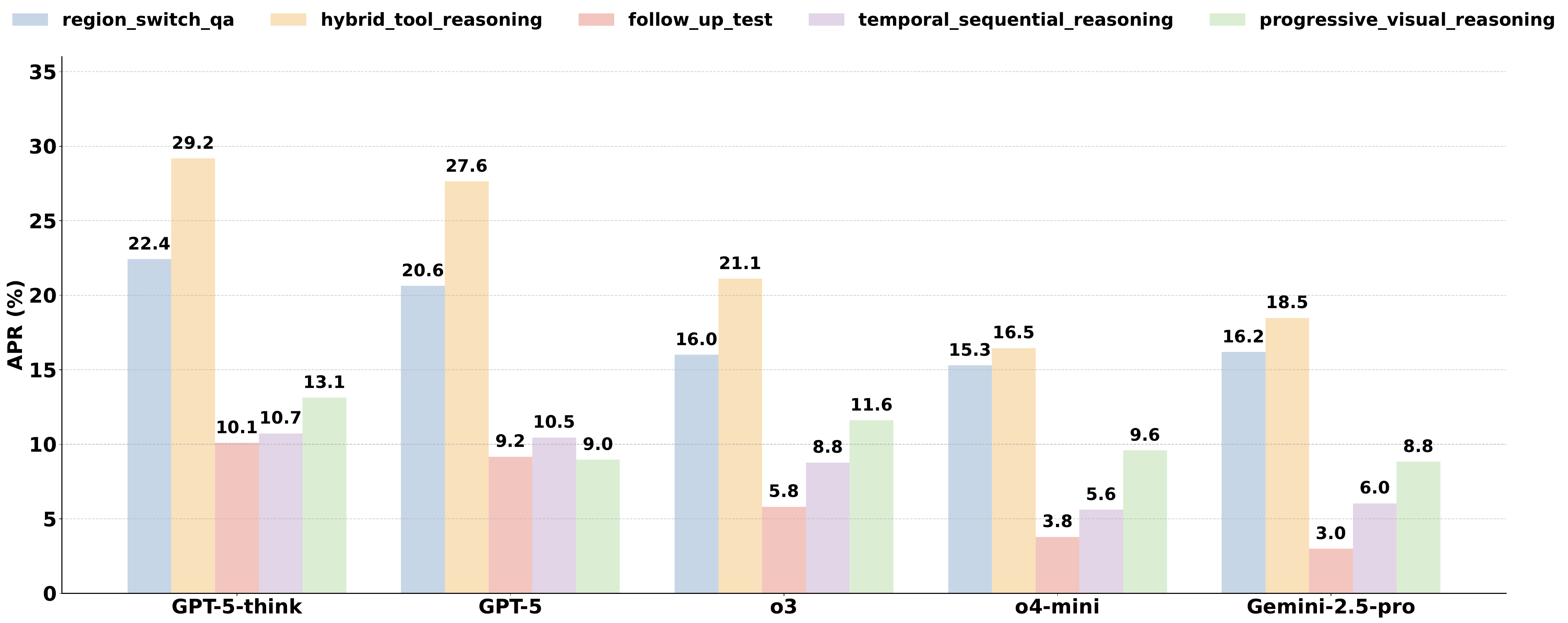}
\end{center}
\caption{{APR across task categories for the top five models on \textsc{VisualToolBench}.} Bars report model performance on each category, with exact APR values labeled above.}
\label{fig:domainwise_ARS}
\end{figure}

We present the overall APR results in Table \ref{main-table}. Figure \ref{fig:domainwise_ARS} represents the APR across task categories for top five performing models on \textsc{VisualToolBench}. We make the following observations.

\textbf{\textsc{VisualToolBench} is highly challenging.} From Table~\ref{main-table}, it is clear that \textsc{VisualToolBench} poses a challenging vision tool-use reasoning benchmark. Specifically, even the best-performing model, GPT-5-think, achieves only an 18.68\% overall pass rate, and 11 out of 16 MLLMs obtain APRs below 10\%. This highlights the limitations of current MLLMs and underscores the substantial room for improvement on visual-reasoning tasks where critical content is not directly accessible and must be extracted through vision tools. 

\textbf{OpenAI models outperform others.} Models from OpenAI, including GPT-5, GPT-5-think, and o3, lead performance with APRs above 13\%, showing a clear margin over competing models. This may be attributed to their specific training for solving \textit{think with images} tasks~\citep{thinking_with_images}. On the other hand, Gemini-2.5-pro also demonstrates relatively strong performance (11.75\%), due to its advanced visual perception capabilities~\citep{comanici2025gemini}.

\textbf{Multi-turn tasks are more difficult than single-turn tasks.} From Table~\ref{main-table} and Figure~\ref{fig:domainwise_ARS}, we see that single-turn tasks (region-switch Q\&A, hybrid-tool reasoning) achieve higher pass rates compared to multi-turn tasks (follow-up test, temporal sequential reasoning, and progressive visual reasoning). This is expected, as multi-turn tasks involve 2 to 5 conversational turns, introducing more opportunities for errors and compounding reasoning challenges.

\subsection{Tool-Use Analysis}
\label{tool-use-analysis}

In this section, we perform tool-call analysis on the evaluated models. First, we quantify tool-use behavior with three descriptive metrics computed from execution traces: \textbf{proactivity}, \textbf{tool-call success rate}, and \textbf{tool-call volume}. {Proactivity} is the fraction of tasks in which at least one tool is invoked, capturing a model’s tendency to integrate tools into its reasoning. {Success rate} measures the proportion of invocations that return a schema-valid, non-empty result, reflecting model's adherence to tool specifications. {Tool-call volume} is the average number of tool calls per task, indicating how heavily a model relies on tools to solve tasks. Precise definitions are provided in Appendix~\ref{app:tool_use_metrics}. 

\begin{figure}[t]
\begin{center}
\includegraphics[width=1.0\linewidth]{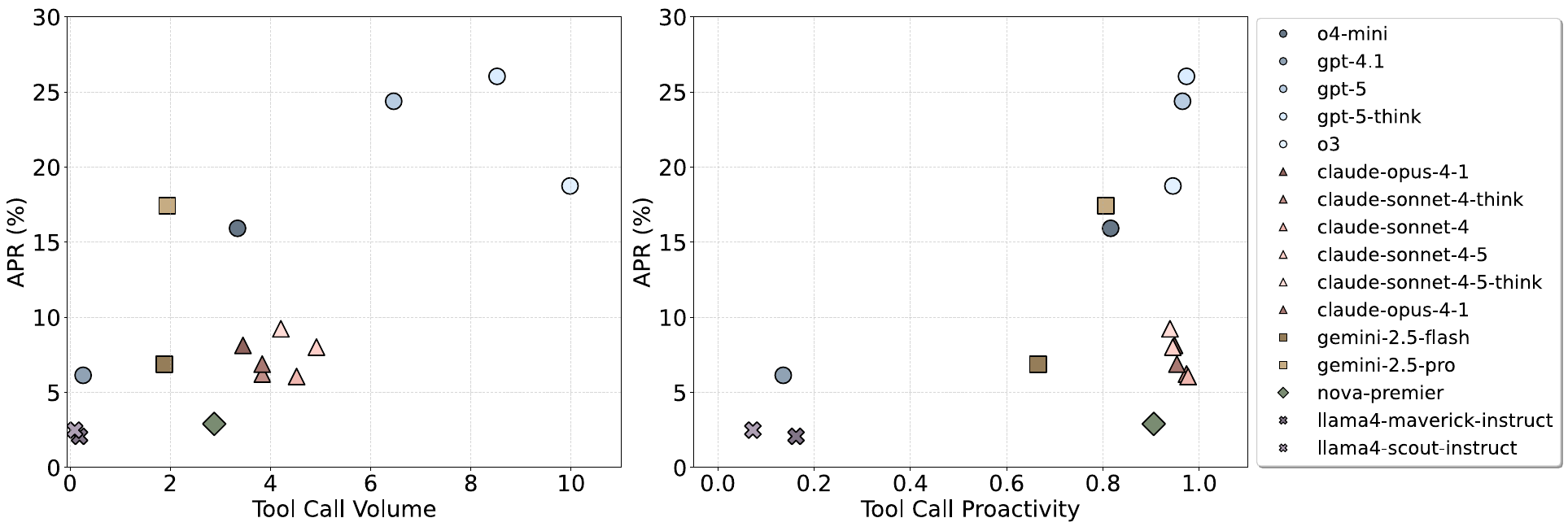}
\end{center}
\caption{Relationship between APR and tool-use behaviors across models. Left: APR versus tool-use proactivity. Right: APR versus average tool call volume per task.}
\label{fig:tool_performance_2D}
\end{figure}

\textbf{Models use tools ineffectively.} Figure~\ref{fig:tool_performance_2D} plots model APR against tool-use proactivity (left) and average tool-call volume per task (right). We observe a clear positive correlation: models that use tools more proactively and with higher call volume tend to achieve better performance. OpenAI models cluster in the upper right, showing both higher APR and stronger tool-use behaviors, while other families (Claude, Gemini, Llama, Nova, Pixtal) remain concentrated in the lower ranges with limited tool use and weaker performance. These findings indicate that most current MLLMs underexploit tool capabilities, and that proactive, consistent tool use is a key driver of success. More detailed tool-call analyses are provided in Appendix~\ref{app:tool_use_analysis}.

\begin{figure}[t]
\begin{center}
\includegraphics[width=1.0\linewidth]{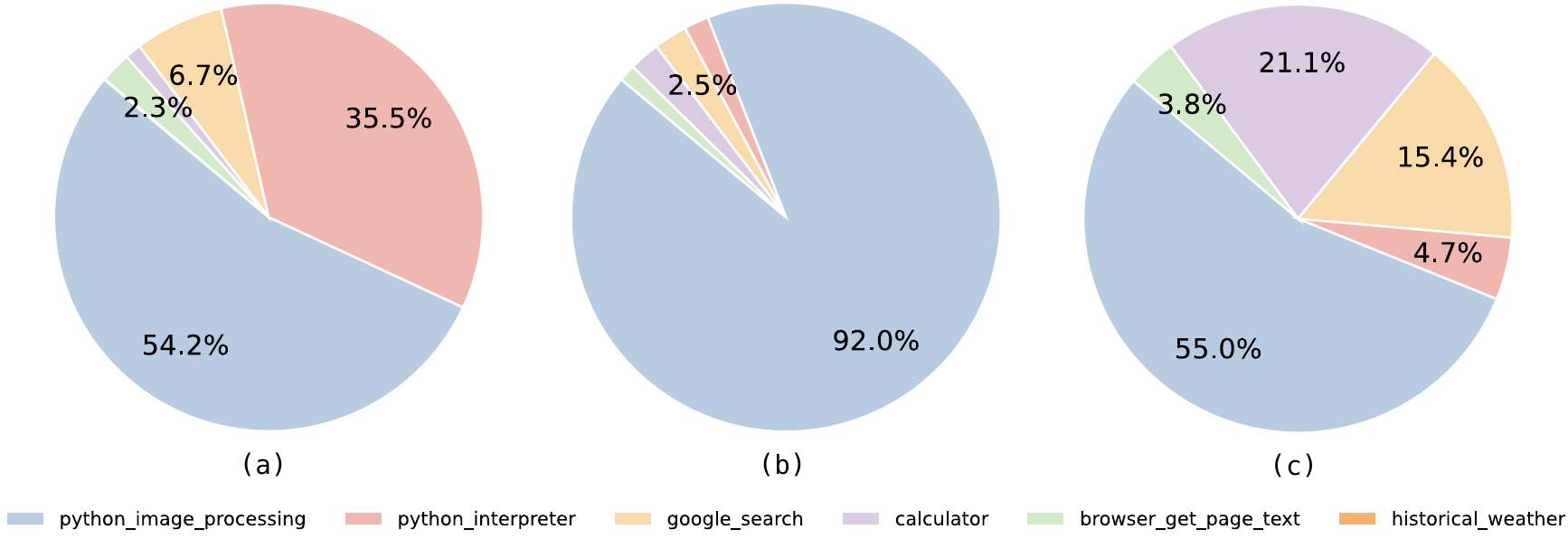}
\end{center}
\caption{Tool-use distribution of top three APR models. Each subplot shows the percentage share of tool calls. Total number of tool calls: (a) {o3}: 6004; (b) {GPT-5}: 3805; (c) {Gemini-2.5-pro}: 1082.}
\label{fig:tool_distribution}
\end{figure}

\textbf{Vision tool is the most called tool}. Figure~\ref{fig:tool_distribution} presents the tool-use distributions of the top three APR models (o3, GPT-5-think, and Gemini-2.5-Pro).
Across all three models, more than 50\% of tool calls are to the \texttt{python\_image\_processing} (92\% for GPT-5), underscoring that image manipulation is the primary operation required to solve tasks in our benchmark. Other tools such as \texttt{python\_interpreter}, \texttt{web\_search}, and \texttt{calculator} are invoked less frequent, reinforcing the inherently \textit{think with images} nature of \textsc{VisualToolBench}.

\begin{figure}[t!]
 \begin{minipage}{0.4\textwidth} 
 \centering
 \fontsize{8.2pt}{\baselineskip}\selectfont 
 \renewcommand\tabcolsep{1.0pt} 
 \renewcommand\arraystretch{0.8} 
 \begin{tabular}{llc}
 \toprule
 \textbf{Model} &\textbf{Error Type} & \textbf{Percentage} \\
 \midrule
 \multirow{4}{*}{{GPT-5}} &visual perception &  \light{71.73\%} \\
 &reasoning & {11.56\%} \\
 &calculation & 2.79\% \\
 &others & 13.92\% \\
 \midrule
 \multirow{4}{*}{{Gemini-2.5-pro}} &visual perception &  \light{78.01\%}\\
 &reasoning &  {12.24\%} \\
 &calculation &  5.74\% \\
 &others & 4.01\% \\
  \midrule
 \multirow{4}{*}{{Claude-opus-4.1}} &visual perception &  \light{82.11\%} \\
 &reasoning &  {9.36\%} \\
 &calculation &  1.84\% \\
 &others &  6.69\% \\
 \bottomrule
 \end{tabular}
 \captionof{table}{Error type statistics.}
 \label{tab:error_analysis}
 \end{minipage} 
 \hfill
 \begin{minipage}{0.6\textwidth}
 \centering
\includegraphics[width=1.0\linewidth]{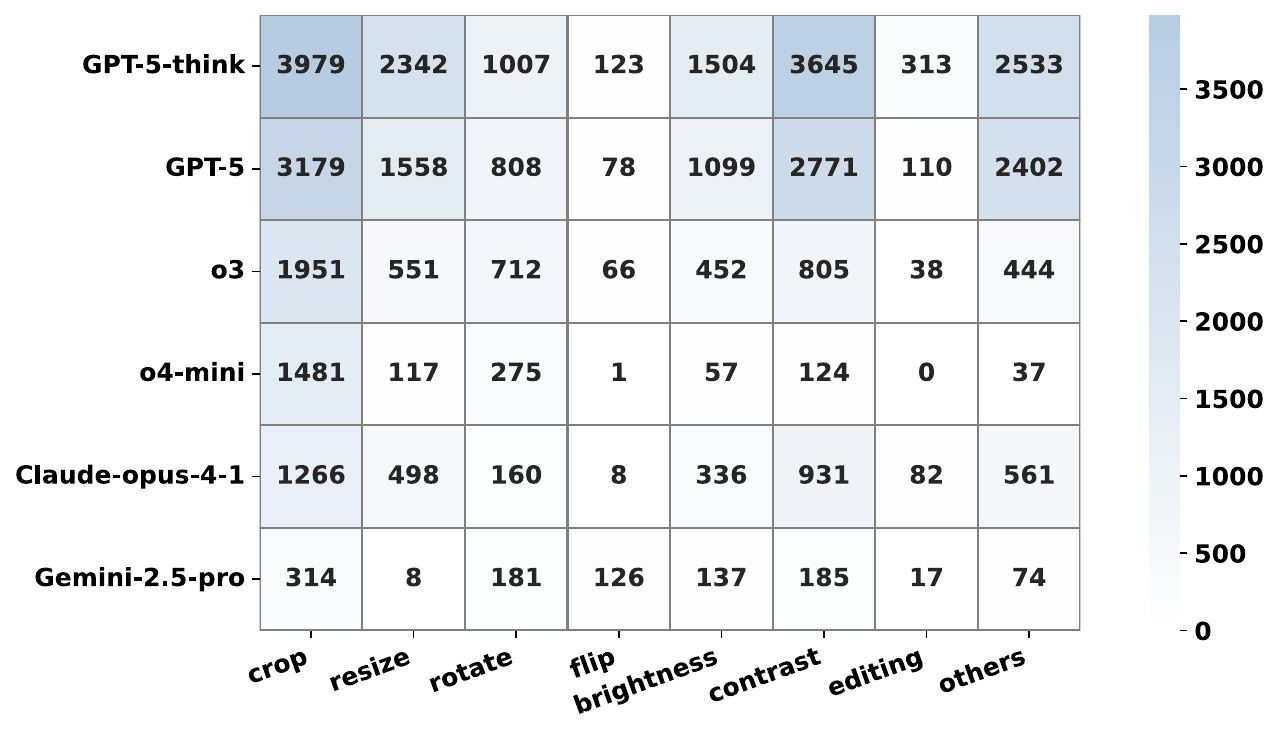}
 \caption{Image manipulation operations counts.}
 \label{fig:image_types}
 \end{minipage}
\end{figure}

\textbf{Models exhibit diverse patterns of image manipulation.} We do not constrain models to a predefined set of image operations; instead, they can invoke arbitrary manipulations via Python through the vision tool. To better understand tool usage in practice, we group the issued calls into eight representative categories: cropping, resizing, rotation, flipping, brightness adjustment, contrast adjustment, editing (e.g., annotation, inpainting, drawing), and others.
Figure~\ref{fig:image_types} reports the distribution of these operations across the top-performing models. GPT-5-think and GPT-5 stand out with both higher volumes and broader diversity of manipulations, reflecting more active exploration of tool capabilities\footnote{Although GPT-5 and GPT-5-think issue fewer tool calls than o3 (see the first column of Table~\ref{tab:tool-call-analysis}), they perform more image manipulations overall as shown in Figure \ref{fig:image_types}. A deeper inspection shows that GPT-5 and GPT-5-think often execute multiple operations within a single vision-tool call, indicating greater tool-use efficiency compared to o3 and contributing to their superior performance. See a demonstration in Appendix~\ref{app:gpt-5-tool-call}.}. o3 and o4-mini also perform frequent manipulations but with narrower operation profiles, while Claude-opus-4-1 and Gemini-2.5-pro show comparatively limited usage.
Illustrative cases and qualitative edge examples are provided in Appendix~\ref{app:image_ops}.

\subsection{Error Analysis}
In this section, we analyze errors for three representative models (GPT-5, Gemini-2.5-pro, and Claude-opus-4.1) using the following four general categories: visual-perception error, reasoning, calculation, and others. Table~\ref{tab:error_analysis} summarizes the distribution of error types. Across all models, visual-perception errors are the most common failure mode. By contrast, calculation mistakes are rare, and reasoning errors occur only occasionally. More discussions are provided in Appendix~\ref{app:error_analysis}.

\subsection{Ablation Study} 
\label{subsec:ablation}

\begin{figure}[h]
\begin{center}
\includegraphics[width=0.9\linewidth]{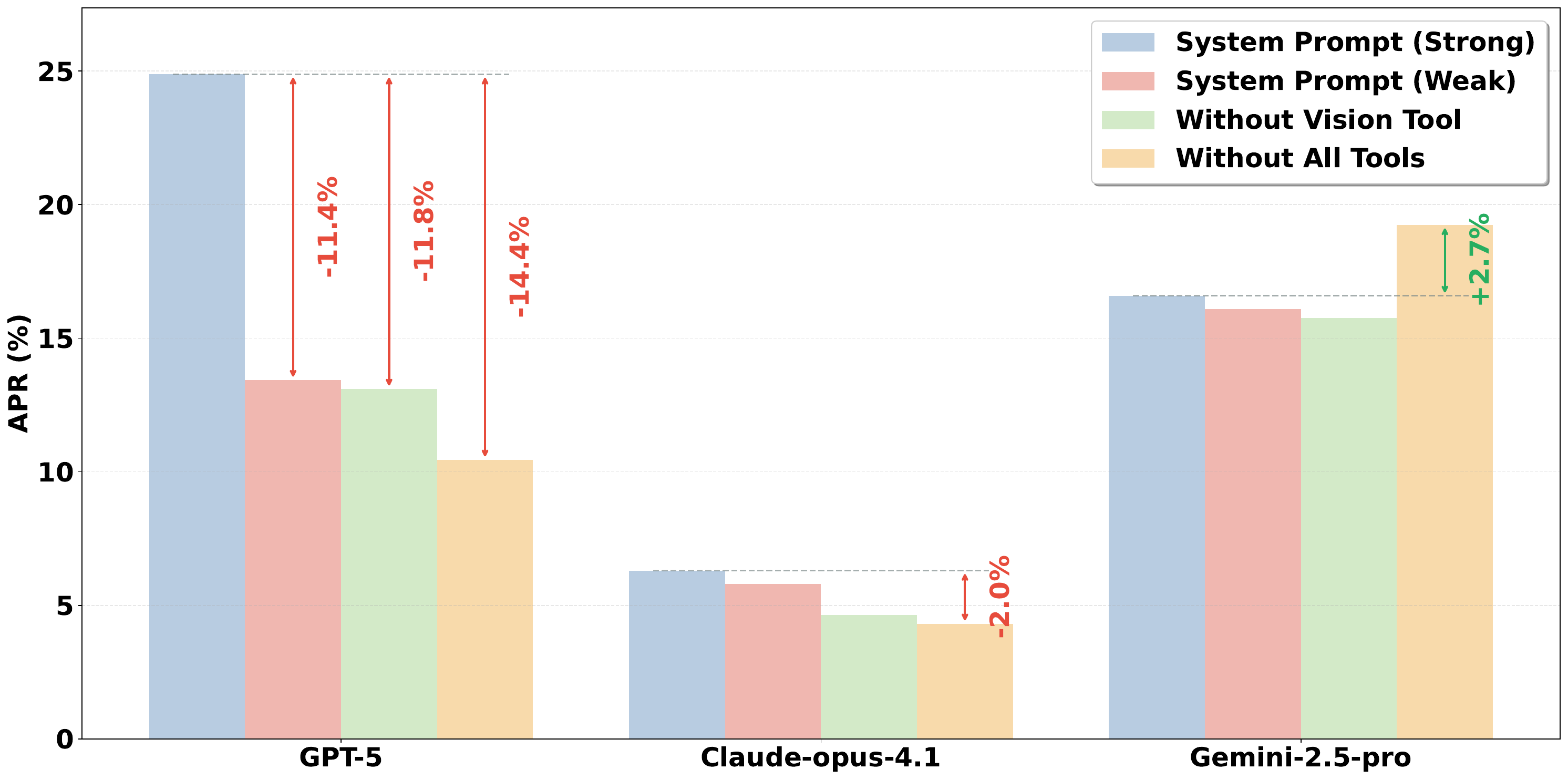}
\end{center}
\caption{Impact of system prompts and tool availability on model performance (APR).
This figure compares the average pass rate (APR) of three representative MLLMs: GPT-5, Claude-opus-4.1, and Gemini-2.5-pro, under four evaluation settings: strong system prompt, weak system prompt, without vision tool, and without all tools.}
\label{fig:ablation}
\end{figure}

Figure~\ref{fig:ablation} compares APR across four different settings: tool-use with strong and weak system prompts, no vision tool, and no tools. For GPT-5, removing tools or weakening the prompt causes significant performance drops ($\approx11-14$\%), showing clear gains from tool-augmented reasoning. 
Surprisingly, Gemini-2.5-pro performs better \textbf{without tools} ($+2.7$\%), implying the tool-use have counter-effect on its performance. In contrast, GPT-5 benefits from tools, boosting its performance further when enabled. These divergent trends likely reflect training differences: GPT-5 appears reinforced on tool-centric workflows, relying on iterative edits to offset weaker perception, whereas Gemini-2.5-pro, with stronger native vision, was likely exposed to fewer tool-use demonstrations and thus degrades when applying unnecessary tool calls. Despite these differences, all evaluated models achieve only modest APR on \textsc{VisualToolBench}, indicating substantial headroom for improving when and how models use tools.

\section{Related Work}

Most existing multimodal benchmarks remain limited to \textit{think about images}: they focus on passive visual Q\&A without active interactions~\citep{rahmanzadehgervi2024vision, yue2024mmmu, wang2024charxiv, lu2023mathvista, zou2024dynamath, wu2024v, scaleai_vista_2025}, restrict tool use to basic operations like cropping~\citep{wang2024gta, ma2024m}, or evaluate tool-agnostic multi-turn dialogues without essential integration~\citep{liu2024mmdu, yan2025mmcr}; see also~\citep{li2024survey} for a broader overview. 

Recent efforts toward teaching MLLMs to \textit{think with images} include prompting-based methods that leverage language mediation, visual input manipulation, or expert integration~\citep{zeng2022socratic, yang2023mmreact, wu2024mind, liu2025visual}, supervised fine-tuning to enable tool invocation and intrinsic manipulations such as cropping, grounding, and dynamic attention~\citep{liu2023llavaplus, shao2024visual}, and reinforcement learning for adaptive exploration and tool orchestration~\citep{wang2025jigsaw, fan2025grit, huang2025visualtoolagent, zheng2025deepeyes}. Despite these advances, evaluations remain constrained to visually passive benchmarks. \textsc{VisualToolBench} addresses this gap by providing a rigorous testbed for genuine visual intelligence under the \textit{think with images} paradigm. See broader discussion in~\citep{su2025thinking} and more related work in Appendix~\ref{app:related_work}.

\section{Conclusion}
\label{sec:conclusion}

In this work, we introduced \textsc{VisualToolBench}, a benchmark designed to evaluate MLLMs under the \textit{think with image} paradigm. Unlike prior efforts that treat images as static inputs, our benchmark emphasizes active visual manipulation through tool use across diverse domains and task types. Our experiments show that current MLLMs continue to struggle on tasks requiring dynamic visual manipulations. We hope \textsc{VisualToolBench} will drive progress toward models that can more effectively think with images and tackle challenging real-world scenarios.

\newpage

\bibliographystyle{iclr2026_conference}

\newpage

\appendix
\addtocontents{toc}{\protect\setcounter{tocdepth}{3}}

\hypersetup{linkcolor=iris}

\tableofcontents 

\newpage

\section{More on Related Work}
\label{app:related_work}

\subsection{Existing MLLM Benchmarks}
\textbf{Existing Multi-modal Benchmarks.}
Existing visual reasoning benchmarks are typically limited: (1) Passive Visual Q\&A: focusing on "look-and-answer" type of tasks without any active interactions \cite{bi2025verify, rahmanzadehgervi2024vision, yue2024mmmu, wang2024charxiv, lu2023mathvista, zou2024dynamath, wu2024v, scaleai_vista_2025}. (2) Superficial Tool Use: including only basic tools like cropping, failing to test deeper image manipulation \cite{wang2024gta, ma2024m}. (3) Tool-Agnostic Conversations: evaluation multi-turn dialogue but without integrating essential tools \cite{liu2024mmdu, yan2025mmcr}. See \citep{li2024survey} for a more comprehensive survey of multimodal benchmarks.

\subsection{Learning to think with images}
\textbf{Prompting Methods} Prompt-based methods enable LMMs to coordinate predefined visual tools without parameter updates, turning static inputs into actively explorable workspaces via in-context learning. Early work like Socratic Models~\citep{zeng2022socratic}, PromptCap~\citep{hu2022promptcap}, and MM-REACT~\citep{yang2023mmreact} showed that language can mediate collaboration, allowing text-only LLMs to function as visual reasoners by orchestrating vision experts through dialogue or targeted captions. Other approaches manipulate inputs directly: visual prompt engineering (e.g., red circles)~\citep{shtedritski2023does,zhang2025vlm2benchcloserlookvlms}, Visualization-of-Thought~\citep{wu2024mind}, and Visual Abstract Thinking~\citep{liu2025visual} enhance perception by highlighting, abstracting, or structuring images, while ZoomEye~\citep{shen2024zoomeye}, ViCrop~\citep{zhang2025mllms}, Chain-of-Spot~\citep{liu2024chain}, and VisuoThink~\citep{wang2025visuothink} extend this into systematic zooming and multimodal tree search. Finally, specialized experts can be integrated via prompting: Set-of-Mark~\citep{yang2023setofmark} leverages segmentation, DetToolChain~\citep{wu2024dettoolchain} structures detection reasoning, DyFO~\citep{li2025dyfo} uses MCTS for adaptive focus, and Visual Thoughts~\citep{cheng2025visual} frames expert outputs as cached “visual thoughts.” Collectively, these works demonstrate that careful prompting—through language mediation, input manipulation, or expert integration—can significantly enhance multimodal reasoning without retraining.

\textbf{Think with Image via SFT}  
Supervised fine-tuning (SFT) is a primary method for teaching LMMs to use external tools or internal visual skills by training on datasets that demonstrate tool invocation and integration. For external orchestration, models like LLaVA-Plus~\citep{liu2023llavaplus}, TACO~\citep{ma2024tacolearningmultimodalaction}, and VTS-V~\citep{bai2025multi} learn to compose tools (e.g., OCR, calculators) and follow procedural chains of reasoning. For internal manipulation, frameworks such as CogCoM~\citep{qi2024cogcom}, VGR~\citep{wang2025vgr}, and UniVG-R1~\citep{bai2025univg} show how SFT can endow models with intrinsic capabilities like cropping, grounding, or fine-grained perception. Finally, SFT also cultivates dynamic visual attention: Visual CoT~\citep{shao2024visual}, IVM~\citep{zheng2024instruction}, CMM-CoT~\citep{zhang2025cmmcot}, selective revisitation~\citep{chung2025don}, and V$^*$~\citep{wu2024v} demonstrate how training with attentional annotations transforms attention into an active, controllable skill. Across these directions, SFT provides the supervision that converts high-level reasoning into executable visual actions.

\textbf{Think with Image via Reinforcement Learning.} 
Reinforcement learning (RL) advances beyond supervised imitation by enabling models to optimize policies for visual reasoning through interaction and feedback. Foundational studies such as Jigsaw-R1~\citep{wang2025jigsaw}, V-Triune~\citep{ma2025one}, and VisionReasoner~\citep{liu2025visionreasoner} established that RL improves generalization over SFT and supports unified frameworks for diverse perception tasks. Building on this, GRIT~\citep{fan2025grit}, Point-RFT~\citep{ni2025point}, and Seg-Zero~\citep{liu2025seg} demonstrated policies that embed spatial cues (e.g., bounding boxes, positional prompts) into reasoning, forming multimodal chains of thought. RL has also enabled active tool orchestration: VisTA~\citep{huang2025visualtoolagent} learns tool-selection policies, Chain-of-Focus~\citep{zhang2025chain} and ACTIVE-o3~\citep{zhu2025active} develop adaptive zooming and region proposals, while DeepEyes~\citep{zheng2025deepeyes} achieves interleaved multimodal reasoning without SFT. Exploration is further incentivized by Pixel-Reasoner~\citep{su2025pixel}, while VILASR~\citep{wu2025reinforcing} leverages drawing-based reasoning, and OpenThinkIMG~\citep{su2025openthinkimg} introduces the first open-source end-to-end RL framework for invoking diverse external tools. Collectively, these approaches move LMMs from passive viewers to active visual agents.

\section{More on Evaluation Results}
\label{app:evaluation_metrics}

\subsection{More on Evaluation Metrics}
\label{app:detail_rubric_score}

In this section, we provide a detailed description on how to evaluate a model's response based on the rubrics. For each task $i \in \{1,2,\cdots, N\}$ in \textsc{VisualToolBench}, we perform the following steps:
\begin{enumerate}
    \item \textbf{Obtain Model's Final Response.} We generate model's final response with \{prompt, image\} pair while allowing model to use a set of predefined tool sets.
    \item \textbf{Rubric Grading.} For each rubric criterion $j \in \{1,2,\cdots, N_i \}$, we use an LLM to grade whether the rubric criterion is met based on the model's response and the rubric criterion.
    \item \textbf{Weighted Rubric Score for Task $i$}. Then we compute a final score for task $i$ using the following weighted average sum:
    \begin{equation}
        s_i = \frac{\sum_{j=1}^{N_i} \mathbbm{1}_{r_{ij}} w_{ij}}{\sum_{j=1}^{N_i} w_{ij} },
    \end{equation}
    where $w_{ij} \in \{1,2,3,4,5\}$ is the assigned weight for each rubric criterion item, and $\mathbbm{1}_{r_{ij}}$ is an indicator representing whether criterion $j$ is met. 
\end{enumerate}

The final score $S$ for the whole benchmark is then computed as the mean value of each task's score:
\begin{equation}
    S = \frac{1}{N} \sum_{i=1}^N s_i.
\end{equation}

\subsection{Rubrics Weights}

To ensure consistent evaluation, each rubric item in \textsc{VisualToolBench} is assigned a weight $w \in \{1,2,3,4,5\}$ that reflects its relative importance.  
Table~\ref{tab:rubric_weights} outlines the five weight levels, ranging from \emph{incidental} stylistic preferences to \emph{critical} elements that determine overall task validity.  
Task contributors are instructed to assign rubric weights in accordance with these guidelines.

\begin{table}[h]
\centering
\renewcommand{\arraystretch}{1}
\begin{tabular}{c|p{10cm}}
\toprule
\textbf{Rubric Weight} & \textbf{Description} \\
\midrule
{Critical (5)} & {Non-negotiable.} This element must be present for the answer to count as valid. Failing it implies the task has failed, regardless of other strengths. \\
\midrule
{Significant (4)} & {Central to success.} Leaving this out degrades output quality or creates confusion about how the task was solved. \\
\midrule
{Moderate (3)} & {Meaningfully important.} Affects clarity or correctness; its absence weakens the answer but does not make it invalid. \\
\midrule
{Minor (2)} & {Adds polish or completeness but is not essential.} Omitting it slightly lowers quality without breaking the core solution. \\
\midrule
{Incidental (1)} & {Marginally relevant.} A nice-to-have detail or stylistic preference that does not impact whether the model solves the task. \\
\bottomrule
\end{tabular}
\caption{Rubric criteria weights used in \textsc{VisualToolBench}. Higher weights indicate greater importance, with critical rubrics determining task-level success.}
\label{tab:rubric_weights}
\end{table}

\subsection{Average Rubric Score Results}

Table \ref{tab:ars_result} presents the detailed rubric score for both single-turn and multi-turn tasks. It can be seen that APR and ARS are positively correlated, models have higher APR also have higher ARS.

\begin{table}[h]
  \caption{ARS results of the evaluated models across domains (averaged across two independent trials). 
  Domain abbreviations:
  \textbf{Med} (Medicine), \textbf{Fin} (Finance), \textbf{Spt} (Sports), 
  and \textbf{Gen} (Generalist). The best results in each column are
highlighted with a red background, and the second-best results are highlighted in blue.}
  \label{tab:ars_result}
  \centering
   \resizebox{1.0\linewidth}{!}{
    \begin{tabular}{l|c|ccccc|ccccc}
      \toprule                
      \multirow{2}{*}{\textbf{Model}} & \multirow{2}{*}{\textbf{Overall}} & \multicolumn{5}{c|}{\textbf{Single-Turn}} & \multicolumn{5}{c}{\textbf{Multi-Turn}} \\
      \cmidrule(lr){3-7} \cmidrule(lr){8-12}
       & & \textbf{STEM} & \textbf{Med} & \textbf{Fin} & \textbf{Sprt} & \textbf{Gen} & \textbf{STEM} & \textbf{Med} & \textbf{Fin} & \textbf{Sprt} & \textbf{Gen}   \\
    \midrule
    \multicolumn{12}{c}{\textbf{Open-Source Models}} \\
    \midrule
    Llama4-Maverick &0.1963 &0.1875 &0.1581 &0.1182 &0.1562 &0.1524 &0.2849 &0.3060 &0.2030 &0.2060 &0.1938\\
    Llama4-Scout &0.1884 &0.1945 &0.1471 &0.1228 &0.1690 &0.1677 &0.2721 &0.2646 &0.2040 &0.2017 &0.1560\\
    \midrule
    \multicolumn{12}{c}{\textbf{Closed-Source Models}} \\
    \midrule
    GPT-4.1 &0.3304 &0.2782 &0.2976 &0.2167 &0.3347 &0.2772 &0.4275 &0.4286 &0.3062 &0.4229 &0.3168\\
    o3 &0.4087 &0.3839 &0.3990 &0.3285 &0.3653 &0.3542 &0.4654 &\light{0.5272} &0.4219 &0.4541 &0.3900\\
    o4-mini &0.4021 &0.3850 &0.4012 &0.3500 &0.3718 &0.4197 &0.4417 &0.4721 &0.3777 &0.4149 &0.3856\\
    GPT-5 &\high{0.4696} &\light{0.4706} &\light{0.4367} &\light{0.4472} &\high{0.3972} &\high{0.4511} &\high{0.5504} &0.5140 &\light{0.4569} &\high{0.4977} &\light{0.4751}\\
    GPT-5-think &\light{0.4712} &\high{0.4485} &\high{0.4270} &\high{0.3983} &\light{0.4365} &\light{0.4760} &\light{0.5674} &\high{0.5294} &\high{0.4544} &\light{0.5044} &\high{0.4696}\\
    Gemini-2.5-pro &0.4130 &0.4197 &0.3514 &0.3969 &0.3644 &0.3945 &0.4559 &0.4973 &0.3915 &0.4693 &0.3877\\
    Gemini-2.5-flash &0.2837 &0.2271 &0.1716 &0.1740 &0.2177 &0.1901 &0.4173 &0.4544 &0.2924 &0.3941 &0.3065\\
    Claude-sonnet-4 &0.2851 &0.2436 &0.2620 &0.2245 &0.2555 &0.2529 &0.3496 &0.4012 &0.2869 &0.3085 &0.2687\\
    Claude-opus-4.1 &0.3056 &0.2575 &0.2625 &0.2413 &0.2717 &0.2800 &0.3894 &0.4281 &0.2800 &0.3471 &0.3018\\
    Claude-sonnet-4.5 &0.3123 &0.2737 &0.2741 &0.2506 &0.3113 &0.2913 &0.3405 &0.4118 &0.3298 &0.3503 &0.2922\\
    Claude-sonnet-4-think &0.2854 &0.2175 &0.2688 &0.2401 &0.2392 &0.2600 &0.3428 &0.4016 &0.3121 &0.2895 &0.2831\\
    Claude-opus-4.1-think &0.2822 &0.2505 &0.2709 &0.2596 &0.2937 &0.2741 &0.3114 &0.3621 &0.2538 &0.2941 &0.2510\\
    Claude-sonnet-4.5-think &0.3114 &0.2628 &0.2860 &0.2692 &0.2528 &0.2939 &0.3664 &0.3855 &0.3354 &0.3590 &0.3009\\
    Nova-Premier &0.2335 &0.1985 &0.2129 &0.1666 &0.2235 &0.1967 &0.3338 &0.3474 &0.2057 &0.2436 &0.2140\\
    \bottomrule
    \end{tabular}
    }
\end{table}

\subsection{Tool-Use Evaluation Metrics}
\label{app:tool_use_metrics}

We evaluate tool-use behaviors based on execution traces recorded by the evaluation harness.  
Let $\mathcal{T}$ denote the set of tasks ($N = |\mathcal{T}|$). For a given task $i \in \mathcal{T}$, let $\mathcal{T}_i = \{c_{i1}, c_{i2}, \dots\}$ represent the (ordered) multiset of tool invocations initiated by the model.

\paragraph{Tool-Call Proactivity.}  
The fraction of tasks in which the model invoked at least one tool:  
\begin{equation}
\mathrm{Proactivity} \;=\; \frac{\bigl|\{\, i \in \mathcal{T} \; \big| \; |\mathcal{T}_i| > 0 \,\}\bigr|}{|\mathcal{T}|}.
\end{equation}
Higher values indicate more frequent tool integration, though proactivity may reflect either beneficial or redundant calls.

\paragraph{Tool-Call Success Rate.}  
The fraction of valid tool calls across all invocations:  
\begin{equation}
\mathrm{Success\ Rate} \;=\;
\frac{\displaystyle \sum_{i \in \mathcal{T}} \; \sum_{c \in \mathcal{T}_i} \mathbbm{1}[\text{valid}(c)]}
{\displaystyle \sum_{i \in \mathcal{T}} |\mathcal{T}_i| },
\end{equation}
where $\mathbbm{1}[\text{valid}(c)]$ is an indicator function that equals $1$ if tool call $c$ succeeds and $0$ otherwise. We determine validity by inspecting tool outputs: error messages imply $\text{valid}(c)=0$, while all other outputs imply $\text{valid}(c)=1$. This metric measures how reliably a model adheres to tool specifications.

\paragraph{Tool-Call Volume.}  
The average number of tool calls per task:  
\begin{equation}
\mathrm{Volume} \;=\; \frac{1}{N} \sum_{i \in \mathcal{T}} |\mathcal{T}_i|,
\end{equation}
where $N$ is the total number of tasks and $|\mathcal{T}_i|$ denotes the number of tool calls made in task $i$.

\subsection{Tool-Use Analysis}
\label{app:tool_use_analysis}

Table \ref{tab:tool-call-analysis} reports the detailed tool-use metrics for all evaluated models. By combining these results with model performance (APR in Table \ref{main-table} and ARS in Table \ref{tab:ars_result}), we make the following key observations:  
\begin{enumerate}
    \item \textbf{More tool calls do not necessarily translate to better performance.} For instance, {o3} makes the largest number of calls (16,116), yet performs worse than {GPT-5} (10,212 calls) and {GPT-5-think} (13,429 calls).  
    \item \textbf{High proactivity does not guarantee strong results.} Claude models exhibit very high proactivity, yet their performance remains poor, with overall APR below 6.5\% and ARS values under 0.35. 
    \item \textbf{Low proactivity and low call volume generally correlate with poor performance.} For example, the Llama models, {GPT-4o}, and {GPT-4.1} all demonstrate relatively low proactivity and correspondingly weak performance.  
\end{enumerate}

\begin{table}[t]
  \caption{Tool call analysis of the evaluated models. The best results in each column are
highlighted with a red background, and the second-best results are highlighted in blue.}
  \label{tab:tool-call-analysis}
  \centering
   \resizebox{1.0\linewidth}{!}{
    \begin{tabular}{l|c|ccc|ccc}
      \toprule                
      \multirow{2}{*}{\textbf{Model}} & \multirow{2}{*}{\textbf{Total \#}} & \multicolumn{3}{c|}{\textbf{Single-Turn}} & \multicolumn{3}{c}{\textbf{Multi-Turn}} \\
      \cmidrule(lr){3-5} \cmidrule(lr){6-8}
       & & \textbf{Proactivity} & \textbf{Success Rate} & \textbf{Volume} & \textbf{Proactivity} & \textbf{Success Rate} & \textbf{Volume}   \\
    \midrule
    \multicolumn{8}{c}{\textbf{Open-Source Models}} \\
    \midrule
    Llama4-Maverick &315 &0.1625 &0.6609 &0.19 &0.2562 &0.1550 &0.33\\
    Llama4-Scout &110 &0.0729 &0.6140 &0.09 &0.0749 &0.0943 &0.09 \\
    \midrule
    \multicolumn{8}{c}{\textbf{Closed-Source Models}} \\
    \midrule
    GPT-4.1 &535 &0.1359 &0.8038 &0.26 &0.2728 &0.3263 &0.63\\
    o3 &\light{16116} &0.9453 &0.8587 &\light{9.98} &0.9750 &0.3696 &\light{16.80} \\
    o4-mini &5337 &0.8159 & 0.8199 &3.34 &0.8835 &0.3138 &5.53\\
    GPT-5 &10212 &0.9652 &0.8555 &6.46 &{0.9883} &0.3548 &10.53\\
    GPT-5-think &\high{13429} &0.9900 &0.8660 &\high{7.45} &\light{0.9950} &0.3105 &\high{13.79}\\    
    Gemini-2.5-pro &3366 &0.8060 &0.7331 &1.94 &0.9251 &0.3191 &3.66 \\
    Gemini-2.5-flash &2313 &0.6650 &0.8605 &1.87 &0.4809 &0.3144 &1.96  \\
    Claude-sonnet-4 &6941 &\light{0.9768} &0.9252 &4.52 &\high{0.9917} &\high{0.4048} &7.01  \\
    Claude-opus-4-1 &6538 &0.9536 &\light{0.9524} &3.83 &{0.9900} &\light{0.4151} &7.03  \\
    Claude-sonnet-4-5 &7247 &0.9451 &\high{0.9491} &4.92 &{0.9882} &{0.3843} &8.73  \\
    Claude-sonnet-4-think &8704 &\high{0.9735} &{0.9352} &3.84 & 0.7671 &0.3946 &3.91 \\
    Claude-opus-4-1-think &4431 &0.9486 &\light{0.9524} &3.45 &\light{0.9950} &0.3941 &6.72 \\
    Claude-opus-4-5-think &6327 &0.9393 &{0.9423} &4.21 &{0.9774} &0.3844 &7.38 \\
    Nova-premier &5109 &0.9055 &0.5444 &2.88 &0.9734 &0.1823 &5.61\\
    \bottomrule
    \end{tabular}
    }
\end{table}

\subsection{More on Error Analysis}
\label{app:error_analysis}

To better understand model weaknesses, we categorize failure cases in \textsc{VisualToolBench} into four major error types:

\paragraph{Visual Perception Error.}  
Errors arising from a model’s inability to correctly perceive, interpret, or extract relevant information from images. Typical cases include misidentifying objects, overlooking salient regions, or unable to extract key visual content.  

\paragraph{Reasoning Error.}  
Errors caused by flawed logical inference or problem-solving steps in the model’s reasoning process. Typical cases include the model's responses contain invalid intermediate steps, contradictions, or logically inconsistent conclusions.  

\paragraph{Calculation Error.}  
Errors stemming from incorrect arithmetic or symbolic computations. These include mistakes in basic arithmetic, misapplication of formulas, or numerical inaccuracies in intermediate or final answers. 

\paragraph{Other Errors.}  
Residual errors that do not fit the above categories. Examples include incomplete responses, refusals, or hit the maximum tool-calls.

\subsection{LLM-as-Judge vs. Human-as-Judge}
\label{llm_as_judge_alignment}

To enable large-scale evaluation of \textsc{VisualToolBench}, we employ LLMs as automatic judges and compare their assessments against human annotations. Table \ref{tab:llm_judge_alignment} reports the overall alignment rate on a subset of \textsc{VisualToolBench}, as well as breakdowns for objective and subjective rubrics across three judge models. 

\begin{table}[h]
\centering
\caption{Alignment rates of different LLMs when serving as judges.}
\label{tab:llm_judge_alignment}
\begin{tabular}{lccc}
\toprule
\textbf{Judge Model} & \textbf{Overall} & \textbf{Objective Rubrics} & \textbf{Subjective Rubrics} \\
\midrule
o4-mini & 0.8807 & 0.9170 & 0.7396 \\
GPT-4.1 & 0.8818 & 0.8983 & 0.8177 \\
GPT-4o  & 0.8701 & 0.8916 & 0.7865 \\
\bottomrule
\end{tabular}
\end{table}

All models achieve high alignment on objective rubrics, while alignment on subjective rubrics is comparatively lower, reflecting the inherent ambiguity of subjective evaluation. Nevertheless, overall alignment rates remain close to 90\%, underscoring the reliability of LLM-as-judge for our benchmark. We choose to use o4-mini as our judge model for the main experimental study.

\section{More on Experimental Setup}

In this section, we provide more details on our experimental setup.

\subsection{Baseline Models}
In Table \ref{table:models_para}, we list the evaluated models’ endpoints and detailed parameter settings. Whenever possible, we set the temperature to 0. For {o3}, {o4-mini}, {GPT-5}, and {GPT-5-think}, the temperature is set to 1. Since GPT-5 is inherently a reasoning model, we use its default reasoning effort (\texttt{medium}) for GPT-5, and {GPT-5-thinking} corresponds to the \texttt{high} reasoning effort. For {o3} and {o4-mini}, we adopt their default reasoning effort (\texttt{medium}). For Claude’s thinking mode, we set the reasoning budget to 5000 tokens. All other model API hyperparameters are kept at their default settings without further customization.

\label{app:evaluated_models}
\begin{table}[h]
  \caption{Model Endpoints and Hyperparameter Setup}
  \label{table:models_para}
  \centering
  \begin{tabular}{c|l|l} 
    \toprule 
    \textbf{Model Provider} &\textbf{Model Endpoint} &Hyperparameter   \\ 
    \midrule 
    \multirow{6}{*}{{OpenAI}} &o3 &\texttt{reason\_effort = "medium"} \\
    &o4-mini &\texttt{reason\_effort = "medium"} \\
    &gpt-4.1 &\texttt{temperature = 0.0}  \\
    &GPT-5 &\texttt{reason\_effort = "medium"} \\
    &GPT-5-thinking &\texttt{reason\_effort = "high"} \\
    \midrule
    \multirow{4}{*}{{Anthropic}} &claude-sonnet-4-20250514 &\texttt{temperature = 0.0} \\
    &claude-sonnet-4-20250514 (thinking)&\texttt{thinking\_budget = 5000} \\
    &claude-opus-4-1 &\texttt{temperature = 0.0} \\
    &claude-opus-4-1 (thinking)&\texttt{thinking\_budget = 5000} \\
    &claude-sonnet-4-5 &\texttt{temperature = 0.0} \\
    &claude-sonnet-4-5 (thinking)&\texttt{thinking\_budget = 5000} \\
    \midrule
    \multirow{2}{*}{{Google}} &gemini-2.5-pro &\texttt{temperature = 0.0} \\
    &gemini-2.5-flash &\texttt{temperature = 0.0} \\
    \midrule 
    \multirow{2}{*}{{Meta}} &llama4-maverick-instruct &\texttt{temperature = 0.0} \\
    &llama4-scout-instruct &\texttt{temperature = 0.0} \\
    \midrule    
    Amazon &nova-premier-v1:0 &\texttt{temperature = 0.0}  \\
    \bottomrule
  \end{tabular}
\end{table}

\subsection{Prompts}

In our main experimental study, we use a strong system prompt to encourage models to use tools towards solving \textsc{VisualToolBench} tasks. In addition, we design a weaker system prompt for ablation study. The strong and weak system prompts are provided below.

\begin{tcolorbox}[
  enhanced,
  breakable,
  colback=red!70!gray!3!white,
  colframe=red!70!gray!60!white,
  title=System Prompt,
  boxrule=0.5mm,
  width=\textwidth,
  arc=1mm,
  auto outer arc,
  before skip=6pt, after skip=6pt
]

\textbf{System Prompt (Strong)}
\vspace{15pt}

You are a \textit{proactive, tool-empowered} visual-reasoning assistant.  

When user supplies an image and requests to solve a problem that requires visual content that are small, ambiguous, or not centered, you must:

1. Examine the image carefullly and mentally list the visual clues most likely to locate the target object.

2. Proactively use the image-processing tools - such as crop, zoom, or enhance - to isolate and clarify the relevant region.

3. Save each transformed image. The updated image will be appended to the conversation for your reference.

4. Iterate as needed. Call the tools repeatedly until the visual evidence is clear enough to answer the user's request.

5. Double-check your observations. Confirm that the final transformed image supports an accurate, confident response before replying to the user.

6. Use other general-purpose tools if needed to answer the user's question.

7. Please use the tools wisely as you have limited tool calls.

\vspace{15pt}
\textbf{System Prompt (Weak)}
\vspace{15pt}

You are a helpful visual reasoning assistant with access to tools to help you answer the user's question.

\end{tcolorbox}

\subsection{LLM-as-Judge Protocol}

We employ an LLM-as-judge framework for large-scale evaluation. To improve reliability, each judgment considers {one rubric at a time}. All rubrics are {atomic}: each rubric targets a single, verifiable fact or behavior, thereby reducing ambiguity, limiting error propagation, and yielding clearer agreement signals. 

The exact judge prompts are provided below. For each case, the judge receives: (i) the original question, (ii) the gold answer, (iii) a single rubric criterion, and (iv) the model’s answer; it returns a verdict (met or not met) with a brief, evidence-grounded justification.

\begin{tcolorbox}[
  enhanced,
  breakable,
  colback=red!70!gray!3!white,
  colframe=red!70!gray!60!white,
  title=LLM Judge Prompt,
  boxrule=0.5mm,
  width=\textwidth,
  arc=1mm,
  auto outer arc,
  before skip=6pt, after skip=6pt
]

You are an expert evaluator tasked with judging whether a model's answer meets a specific rubric criterion.  
You will be provided with:

- a question

- a golden (reference) answer

- a rubric criterion

- the model's answer  

\vspace{3pt}

Your task is to decide if the model's answer **meets** or **does not meet** the given rubric criterion, referencing the golden answer only as needed.
\vspace{3pt}

\#\#\# Inputs:

**Question:** {question}  

**Golden Answer:** {golden\_answer}  

**Rubric Criterion:** {rubric\_criteria}  

**Model Answer:** {model\_answer}  
\vspace{3pt}

\#\#\# Important Notes:

- The model's answer does not need to be correct to meet the criterion if correctness is not required.  

*Example:* If the rubric is "The model should show its reasoning process to answer the question," the answer can be incorrect but still meet the rubric if model's reasoning process is present.  

- For writing style or presentation rubrics, apply leniency.  

*Example:* If the rubric asks for conciseness, answers that are slightly longer than the golden answer but still reasonably length should be considered as meeting the rubric.

- The model's answer may satisfy the rubric implicitly without explicitly mentioning the exact term. This should still be considered as meeting the criterion if model's answer is reasonable and makes sense.  

*Example:* If the rubric is "The model should demonstrate understanding of photosynthesis," and the model states "Plants make their own food using sunlight," without explicitly mentioning the term "photosynthesis," it still meets the criterion.
\vspace{3pt}

\#\#\# Output Format:

Return your judgment in the following JSON format:
\begin{Verbatim}[fontsize=\footnotesize, breaklines, breakanywhere]
{
    "explanation": "Brief explanation of your judgment",
    "judge_result": "Met" or "Not Met"
}
\end{Verbatim}

\end{tcolorbox}

\section{More on Tools Analysis}
\label{app:tool_description}
In this section, we provide more details on the tools supported by \textsc{VisualToolBench} and more examples on Tool-use of the evaluated models.

\subsection{Tool Description}

Table \ref{tab:tools} provides tool description supported by \textsc{VisualToolBench}.

\begin{table}[t]
\small
\centering
\renewcommand{\arraystretch}{1}
\begin{tabular}{p{4.7cm}|p{8.4cm}}
\toprule
\textbf{Tool} & \textbf{Functionality} \\
\midrule
\texttt{python\_image\_processing} & Vision-specific manipulation, including image cropping, editing, rotation, brightness/contrast adjustment, and enhancement. Enables iterative refinement of visual inputs. \\
\midrule
\texttt{python\_interpreter} & General code execution. \\
\midrule
\texttt{web\_search} & Open-domain information retrieval from the web. \\
\midrule
\texttt{browser-get-page-text} & Extraction of textual content from online sources. \\
\midrule
\texttt{historical\_weather} & Weather records for temporal and geographic look-ups. \\
\midrule
\texttt{calculator} & Arithmetic operations for quick computations. \\
\bottomrule
\end{tabular}
\caption{Supported tools supported in \textsc{VisualToolBench}.}
\label{tab:tools}
\end{table}

\subsection{Image Manipulation Operations}
\label{app:image_ops}

In this section, we present the representative o3 generated python code snippets, where the model invokes the \texttt{python\_image\_processing} tool to perform diverse image manipulations in support of problem solving. Lines that execute the manipulations are highlighted in yellow.

\begin{tcolorbox}[
  enhanced,
  breakable,
  colback=orange!70!gray!3!white,
  colframe=orange!70!gray!60!black,
  title={Image Operations: Crop and Resize},
  boxrule=0.5mm,
  width=\textwidth,
  arc=1mm,
  auto outer arc,
  before skip=6pt, after skip=6pt
]

\textbf{Task ID: 68658a711603983919432615}
\vspace{5pt}

{\color{amaranth}
- line 9-15: image cropping

- line 17-18: image resize}

\begin{lstlisting}[style=pythonstyle]
from PIL import Image
import os
import numpy as np
import cv2

img_path = '/mnt/efs/.../vision_tool_use/Kb2JElJ87npHXeG.jpg'
img = Image.open(img_path)

# Crop bottom middle area where Dixit might be
w, h = img.size
left = int(w * 0.35)
top = int(h * 0.45)
right = int(w * 0.65)
bottom = int(h * 0.8)
crop = img.crop((left, top, right, bottom))

# Enlarge cropped region
crop_large = crop.resize((crop.size[0]*2, crop.size[1]*2))

# Save
os.makedirs('eval_results_single_turn/...', exist_ok=True)
crop_large.save('eval_results_single_turn/...', 'PNG')
\end{lstlisting}

\begin{center}
    \includegraphics[width=1.0\textwidth]{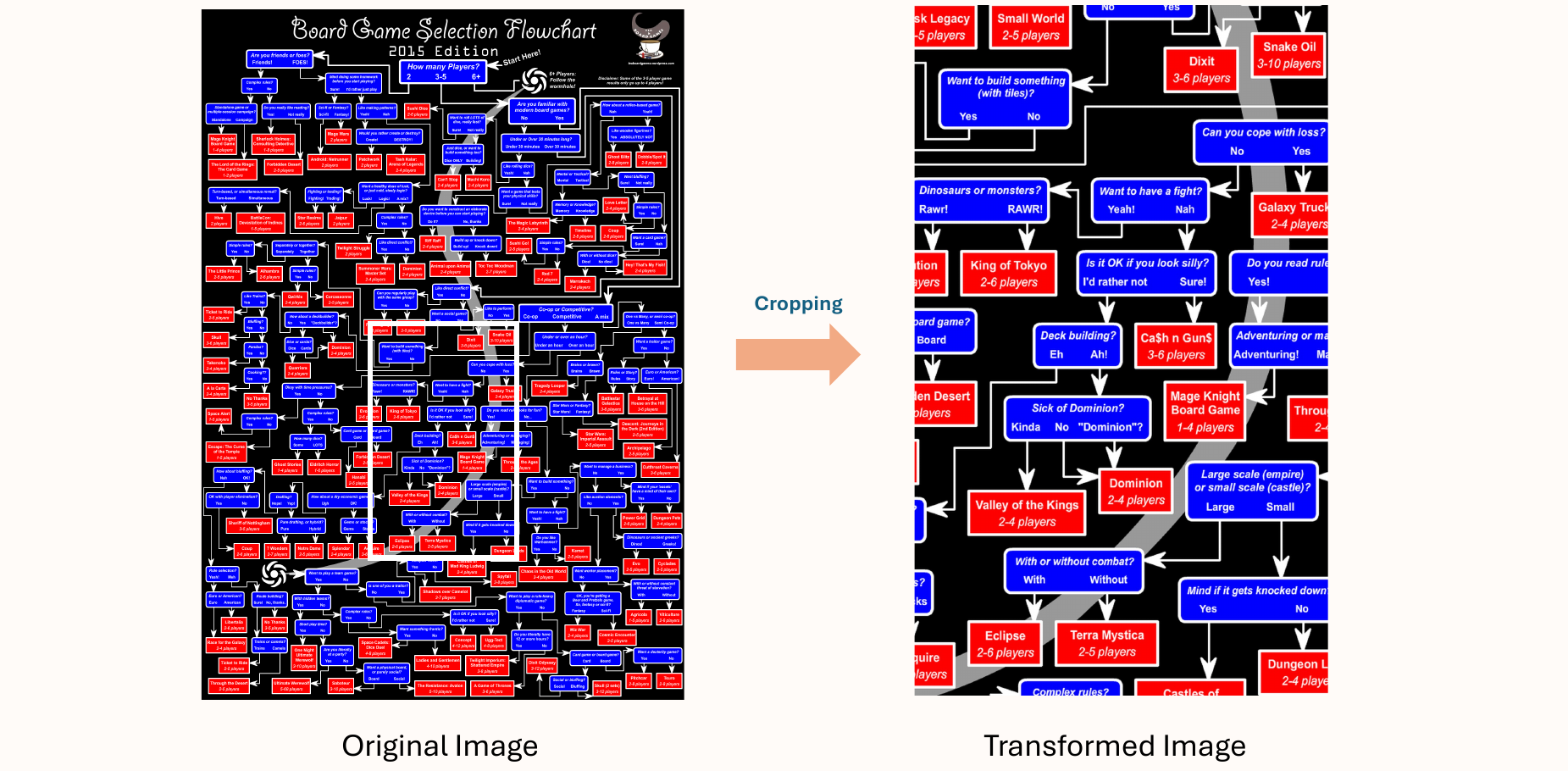}
\end{center}

\end{tcolorbox}

\begin{tcolorbox}[
  enhanced,
  breakable,
  colback=orange!70!gray!3!white,
  colframe=orange!70!gray!60!black,
  title={Image Operations: Rotation, Brightness and Contrast Adjustment},
  boxrule=0.5mm,
  width=\textwidth,
  arc=1mm,
  auto outer arc,
  before skip=6pt, after skip=6pt
]

\textbf{Task ID: 68658a711603983919432613}
\vspace{5pt}

{\color{amaranth}
- line 5-6: image rotation

- line 7-9: brightness adjustment

- line 10-12: contrast adjustment}
\begin{lstlisting}[style=pythonstyle]
from PIL import Image, ImageEnhance
import os, numpy as np
img_path = '/mnt/.../vision_tool_use/PtZEjeYmkRmW-Lf.jpg'
img = Image.open(img_path)
# rotate 180 degrees
img_rot = img.rotate(180, expand=True)
# enhance brightness
enhancer = ImageEnhance.Brightness(img_rot)
img_enh = enhancer.enhance(1.8)
# increase contrast
enhancer_c = ImageEnhance.Contrast(img_enh)
img_enh2 = enhancer_c.enhance(1.5)
# save
save_dir = 'eval_results_single_turn/...'
os.makedirs(save_dir, exist_ok=True)
save_path = os.path.join(save_dir, 'transformed_image_0.png')
img_enh2.save(save_path, 'PNG')
\end{lstlisting}

\begin{center}
    \includegraphics[width=1.0\textwidth]{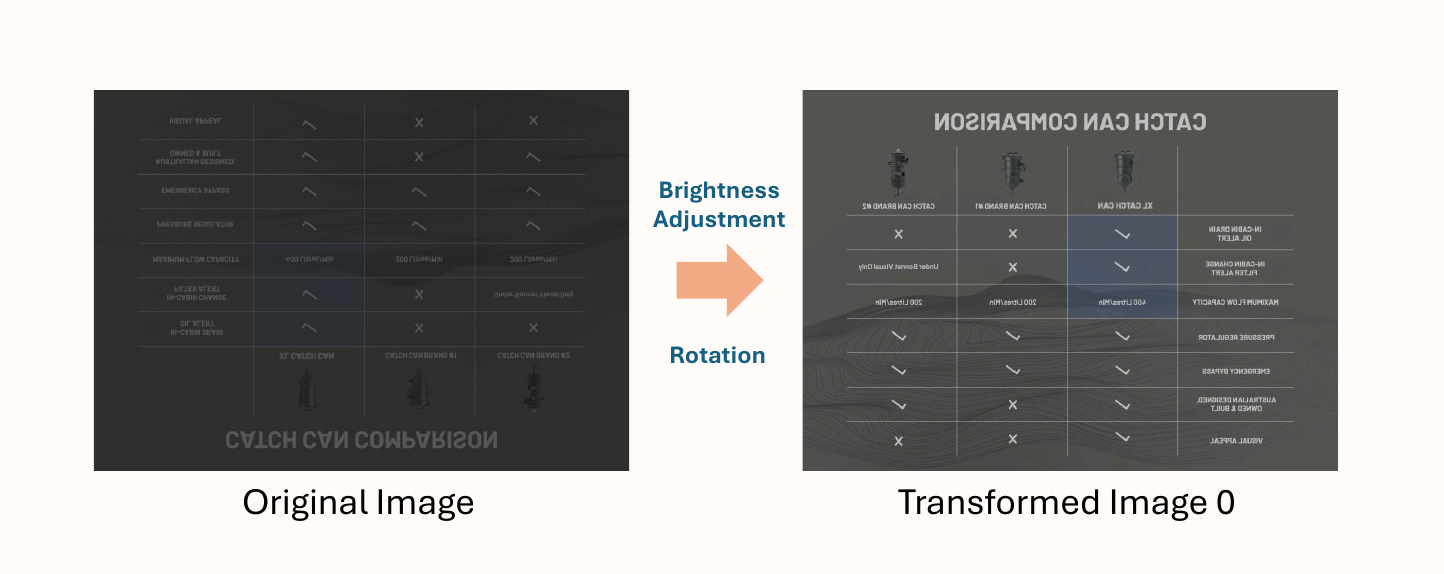}
\end{center}

\end{tcolorbox}

\begin{tcolorbox}[
  enhanced,
  breakable,
  colback=orange!70!gray!3!white,
  colframe=orange!70!gray!60!black,
  title={Image Operations: Flip},
  boxrule=0.5mm,
  width=\textwidth,
  arc=1mm,
  auto outer arc,
  before skip=6pt, after skip=6pt
]

\textbf{Task ID: 68658a711603983919432613}
\vspace{5pt}

{\color{amaranth}- line 4-5: image flip}
\begin{lstlisting}[style=pythonstyle]
from PIL import Image
import os, numpy as np
img = Image.open('/.../transformed_image_0.png')
# flip horizontally to correct mirroring
img_flipped = img.transpose(Image.FLIP\_LEFT\_RIGHT)
# maybe crop to table area? Keep full for readability
save_dir='eval_results_single_turn/...'
img_flipped.save(save_dir,'PNG')
\end{lstlisting}

\begin{center}
    \includegraphics[width=1.0\textwidth]{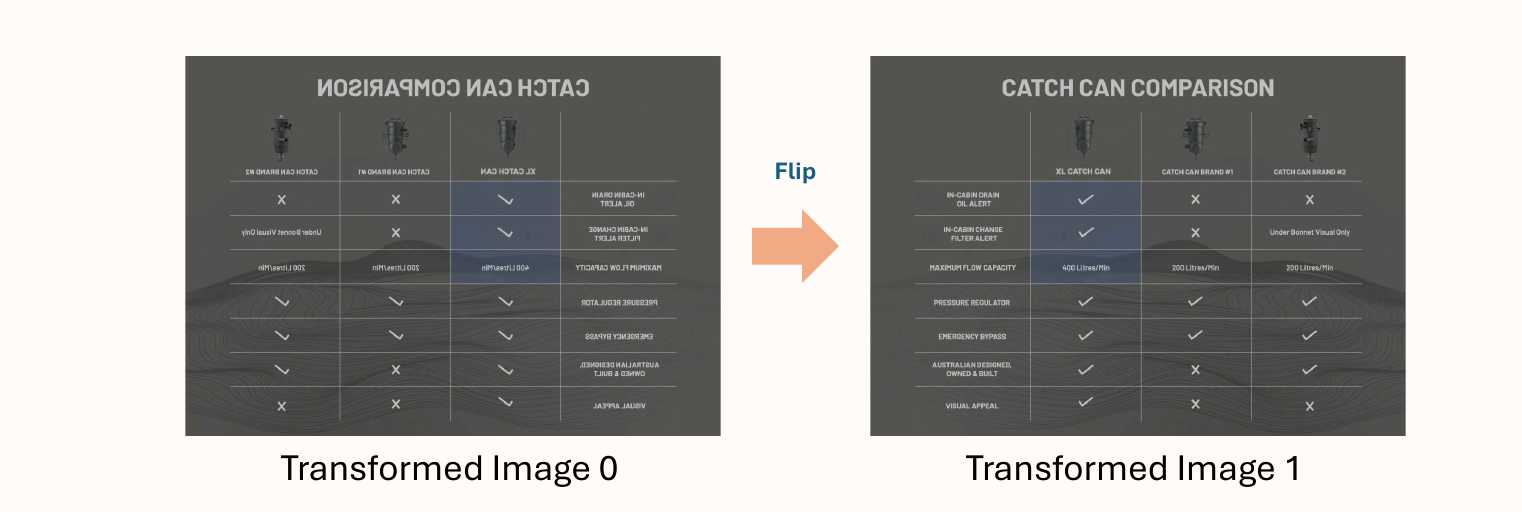}
\end{center}

\end{tcolorbox}

\begin{tcolorbox}[
  enhanced,
  breakable,
  colback=orange!70!gray!3!white,
  colframe=orange!70!gray!60!black,
  title={Image Operations: Editing, Others},
  boxrule=0.5mm,
  width=\textwidth,
  arc=1mm,
  auto outer arc,
  before skip=6pt, after skip=6pt
]

\textbf{Task ID: 68757ac94c044bac23c9bd8b}
\vspace{5pt}

{\color{amaranth}- line 23-26: image editing}
\begin{lstlisting}[style=pythonstyle]
from PIL import Image
import numpy as np
import cv2
import os
# Load image
img_path='/mnt/.../SZJPcfAUQgyEVGe.jpg'
img=cv2.imread(img_path)
# Convert to HSV
hsv=cv2.cvtColor(img, cv2.COLOR_BGR2HSV)
# Define green range (for tree foliage), adjust
lower=np.array([35,40,40])
upper=np.array([85,255,255])
mask=cv2.inRange(hsv, lower, upper)
# Morphological operations to remove noise
kernel=np.ones((5,5),np.uint8)
mask=cv2.morphologyEx(mask, cv2.MORPH_OPEN, kernel)
# Find contours
contours,_=cv2.findContours(mask, cv2.RETR_EXTERNAL,
cv2.CHAIN_APPROX_SIMPLE)
# Filter contours by area to avoid tiny bits
tree_contours=[cnt for cnt in contours if cv2.contourArea(cnt)>500]
print('Totalcontourslarge', len(tree_contours))
# Draw bounding boxes to visualize
for cnt in tree_contours:
  x,y,w,h=cv2.boundingRect(cnt)
  cv2.rectangle(img,(x,y),(x+w,y+h),(0,0,255),2)
# Save transformed
out_dir='/...'
os.makedirs(out_dir, exist_ok=True)
Image.fromarray(cv2.cvtColor(img, cv2.COLOR_BGR2RGB)).
save(out_dir+'/transformed_image_0.png','PNG')
\end{lstlisting}

\begin{center}
    \includegraphics[width=1.0\textwidth]{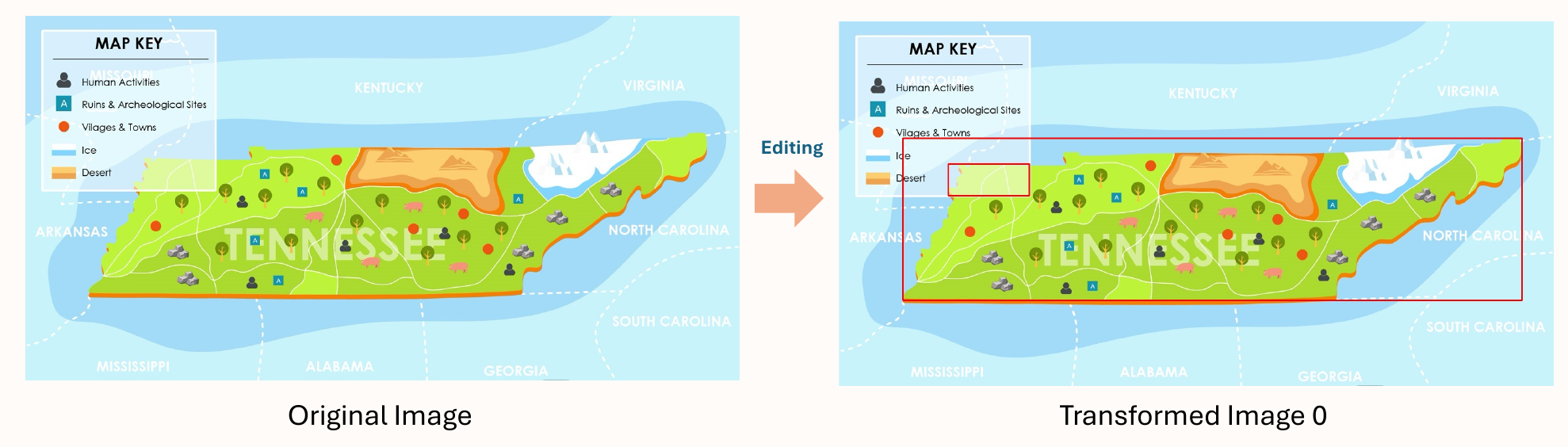}
\end{center}

\end{tcolorbox}

\subsection{Vision Tool Call for GPT-5}
\label{app:gpt-5-tool-call}

From the first column of Table~\ref{tab:tool-call-analysis} and Figure~\ref{fig:image_types}, we observe that although GPT-5 and GPT-5-think invoke the vision tool less frequently than o3, they perform more image manipulations overall. A closer inspection shows that GPT-5 and GPT-5-think often execute multiple image-processing operations within a single vision-tool call, thereby achieving higher tool-call efficiency than o3. We provide two demonstration examples below.

\begin{tcolorbox}[
  enhanced,
  breakable,
  colback=orange!70!gray!3!white,
  colframe=orange!70!gray!60!black,
  title={GPT-5 Vision Tool Call Output},
  boxrule=0.5mm,
  width=\textwidth,
  arc=1mm,
  auto outer arc,
  before skip=6pt, after skip=6pt
]

\textbf{Task ID: 68658a711603983919432613}
\vspace{5pt}

{\color{amaranth}-line 12-14, 21-26: saved three plus transformed images}

\begin{lstlisting}[style=pythonstyle]
from PIL import Image, ImageEnhance, ImageFilter
import os
base='eval_results_single_turn/.../transformed_image_0.png'
img=Image.open(base)
# Let's crop central table area
w,h=img.size
# Crop left column, middle, right ... for readability
# First overall crop removing margins
crop=img.crop((int(w*0.06), ..., int(h*0.92)))
# enhance
crop=ImageEnhance.Contrast(crop).enhance(1.4)
#Save
out_dir=os.path.dirname(base)
crop.save(out_dir+'/transformed_image_1.png','PNG')
# Crop each column region roughly
# Determine thirds
cw=crop.size[0]
ch=crop.size[1]
col_width=cw//3
cols=[]
for i in range(3):
  c=crop.crop((i*col_width,0,(i+1)*col_width,ch)) 
  c=ImageEnhance.Brightness(c).enhance(1.2)
  c=c.filter(ImageFilter.UnsharpMask(radius=1.5,percent=180,threshold=2))
  c.save(out_dir+f'/transformed_image_{2+i}.png','PNG')
  print('saved 3+ images')
\end{lstlisting}

\end{tcolorbox}

\begin{tcolorbox}[
  enhanced,
  breakable,
  colback=orange!70!gray!3!white,
  colframe=orange!70!gray!60!black,
  title={GPT-5-think Vision Tool Call Output},
  boxrule=0.5mm,
  width=\textwidth,
  arc=1mm,
  auto outer arc,
  before skip=6pt, after skip=6pt
]

\textbf{Task ID: 68658a711603983919432616}
\vspace{5pt}

{\color{amaranth}-line 24, 27, 32, 35, 38, 41: saved six transformed images in one tool call}
\begin{lstlisting}[style=pythonstyle]
from PIL import Image, ImageEnhance, ImageFilter, ImageOps
import numpy as np
import os

# Load the source image
src_path = '/mnt/efs/.../ymHX_cj7Vg3lDk8.jpg'
im = Image.open(src_path)
# Step 1: Auto-orient and slightly crop borders
im = ImageOps.exif_transpose(im)
# Step 2: Enhance brightness and contrast
bright = ImageEnhance.Brightness(im).enhance(1.8)
contrast = ImageEnhance.Contrast(bright).enhance(2.0)
# Step 3: Convert to grayscale for clarity
gray = ImageOps.grayscale(contrast)
# Step 4: Increase sharpness
sharp = ImageEnhance.Sharpness(gray).enhance(2.5)
# Step 5: Apply slight denoise via median filter, then unsharp mask
median = sharp.filter(ImageFilter.MedianFilter(size=3))
unsharp = median.filter(ImageFilter.UnsharpMask(radius=2, percent=200, threshold=3))

# Save the enhanced full image
save_dir = 'eval_results_single_turn/...'
os.makedirs(save_dir, exist_ok=True)
unsharp.save(os.path.join(save_dir, 'transformed_image_0.png'), 'PNG')
# Also create a higher-contrast inverted version which sometimes helps
inv = ImageOps.invert(unsharp)
inv.save(os.path.join(save_dir, 'transformed_image_1.png'), 'PNG')
# Crop likely totals area: lower half of the long receipt\nw, h = im.size
# Long receipt is on right; crop right area
right_x0 = int(w*0.35)
right = unsharp.crop((right_x0, 0, w, h))
right.save(os.path.join(save_dir, 'transformed_image_2.png'), 'PNG')
# Crop bottom of right receipt where totals usually appear
bottom_right = unsharp.crop((right_x0, int(h*0.55), w, h))
bottom_right.save(os.path.join(save_dir, 'transformed_image_3.png'), 'PNG')
# Crop mid section (items + discounts)
mid_right = unsharp.crop((right_x0, int(h*0.30), w, int(h*0.60)))
mid_right.save(os.path.join(save_dir, 'transformed_image_4.png'), 'PNG')
# Crop the smaller left slip (maybe PWD details)
left = unsharp.crop((0, int(h*0.1), int(w*0.38), int(h*0.7)))
left.save(os.path.join(save_dir, 'transformed_image_5.png'), 'PNG')
print('Saved enhanced and cropped images.')

\end{lstlisting}

\end{tcolorbox}

\subsection{Tool APIs}
In this section, we provide more details on the tool API description.
\begin{tcolorbox}[
  enhanced,
  breakable,
  colback=blue!70!gray!3!white,
  colframe=blue!70!gray!60!black,
  title=Tool Description: python\_image\_processing,
  boxrule=0.5mm,
  width=\textwidth,
  arc=1mm,
  auto outer arc,
  before skip=6pt, after skip=6pt
]
\begin{Verbatim}[fontsize=\footnotesize, breaklines, breakanywhere]
{
    "type": "function",
    "function": {
        "name": "python_image_processing",
        "description": (
            "Generate arbitrary Python code for image manipulation and save the transformed image as PNG.\n"
            f"-Read one source image (your choice) from the working-directory file list: {image_list}.\n"
            f"-Perform any image processing with PIL, NumPy, or OpenCV. You cannot use matplotlib to show the image.\n"
            f"-You **must save** the transformed image as PNG to {processed_image_save_path} using the filename pattern "
            "\"transformed_image_i.png\", where the counter **i starts at 0 and increments on each invocation** "
            "so files are never overwritten. Example:\n"
            f"    img.save(f\"{processed_image_save_path}/transformed_image_{{i}}.png\", \"PNG\")\n"
        ),
        "parameters": {
            "type": "object",
            "properties": {
                "code": {
                    "type": "string",
                    "description": "Python code to run.",
                    "minLength": 1,
                    "maxLength": 5000
                }
            },
            "required": ["code"]
        }
    }
}
\end{Verbatim}
\end{tcolorbox}

\begin{tcolorbox}[
  enhanced,
  breakable,
  colback=blue!70!gray!3!white,
  colframe=blue!70!gray!60!black,
  title=Tool Description: python\_processing,
  boxrule=0.5mm,
  width=\textwidth,
  arc=1mm,
  auto outer arc,
  before skip=6pt, after skip=6pt
]
\begin{Verbatim}[fontsize=\footnotesize, breaklines, breakanywhere]
{
    "type": "function",
    "function": {
        "name": "python_interpreter",
        "description": (
            "General-purpose Python interpreter. Run arbitrary Python code and capture stdout via print(). "
            "Any exceptions are returned in stderr.\n\n"
            "Pre-installed packages:\n"
            "  • numpy\n"
            "  • pandas\n"
            "  • requests\n"
            "  • scipy\n"
            "  • scikit-learn\n"
            "  • simpy\n"
            "  • tabulate\n"
            "  • beautifulsoup4\n"
            "  • yfinance"
        ),
        "parameters": {
            "type": "object",
            "properties": {
                "code": {
                    "type": "string",
                    "description": "Python code to run.",
                    "minLength": 1,
                    "maxLength": 5000
                }
            },
            "required": ["code"]
        }
    }
}
\end{Verbatim}
\end{tcolorbox}

\begin{tcolorbox}[
  enhanced,
  breakable,
  colback=blue!70!gray!3!white,
  colframe=blue!70!gray!60!black,
  title=Tool Description: web\_search,
  boxrule=0.5mm,
  width=\textwidth,
  arc=1mm,
  auto outer arc,
  before skip=6pt, after skip=6pt
]
\begin{Verbatim}[fontsize=\footnotesize, breaklines, breakanywhere]
{
    "type": "function",
    "function": {
        "name": "web_search",
        "description": (
            "Perform a Google search and return relevant results. "
            "Useful for finding current information, news, or facts about topics."
        ),
        "parameters": {
            "type": "object",
            "properties": {
                "query": {
                    "type": "string",
                    "description": "The search query to look up"
                },
                "num_results": {
                    "type": "integer",
                    "description": "Number of results to return (1-10)",
                    "default": 5
                }
            },
            "required": ["query"]
        }
    }
}
\end{Verbatim}
\end{tcolorbox}

\begin{tcolorbox}[
  enhanced,
  breakable,
  colback=blue!70!gray!3!white,
  colframe=blue!70!gray!60!black,
  title=Tool Description: browser\_get\_page\_text,
  boxrule=0.5mm,
  width=\textwidth,
  arc=1mm,
  auto outer arc,
  before skip=6pt, after skip=6pt
]
\begin{Verbatim}[fontsize=\footnotesize, breaklines, breakanywhere]
{
    "type": "function",
    "function": {
        "name": "browser_get_page_text",
        "description": (
            "Fetch a web page and extract its text content. "
            "Useful for reading articles, documentation, or any web page content."
        ),
        "parameters": {
            "type": "object",
            "properties": {
                "url": {
                    "type": "string",
                    "description": "The URL of the web page to fetch"
                }
            },
            "required": ["url"]
        }
    }
}
\end{Verbatim}
\end{tcolorbox}

\begin{tcolorbox}[
  enhanced,
  breakable,
  colback=blue!70!gray!3!white,
  colframe=blue!70!gray!60!black,
  title=Tool Description: historical\_weather,
  boxrule=0.5mm,
  width=\textwidth,
  arc=1mm,
  auto outer arc,
  before skip=6pt, after skip=6pt
]
\begin{Verbatim}[fontsize=\footnotesize, breaklines, breakanywhere]
{
    "type": "function",
    "function": {
        "name": "historical_weather",
        "description": (
            "Get historical weather data for a specific location and date. "
            "Useful for analyzing past weather patterns or events."
        ),
        "parameters": {
            "type": "object",
            "properties": {
                "location": {
                    "type": "string",
                    "description": "City name or coordinates (e.g., 'New York, NY' or '40.7128,-74.0060')"
                },
                "date": {
                    "type": "string",
                    "description": "Date in YYYY-MM-DD format"
                }
            },
            "required": ["location", "date"]
        }
    }
}
\end{Verbatim}
\end{tcolorbox}

\begin{tcolorbox}[
  enhanced,
  breakable,
  colback=blue!70!gray!3!white,
  colframe=blue!70!gray!60!black,
  title=Tool Description: calculator,
  boxrule=0.5mm,
  width=\textwidth,
  arc=1mm,
  auto outer arc,
  before skip=6pt, after skip=6pt
]
\begin{Verbatim}[fontsize=\footnotesize, breaklines, breakanywhere]
{
    "type": "function",
    "function": {
        "name": "calculator",
        "description": "A calculator tool that can perform basic arithmetic operations including +, -, *, /, %, ^, sqrt, sin, cos, tan, log, exp, and parentheses.",
        "parameters": {
            "type": "object",
            "properties": {
                "expression": {"type": "string", "description": "The expression to evaluate, e.g. \"2 * 3.14 * 5\"."}
            },
            "required": ["expression"]
        }
    }
}
\end{Verbatim}
\end{tcolorbox}

\section{More Benchmark Examples of \textsc{VisualToolBench}}
\label{app:demo_examples}

\begin{figure}[h]
\begin{center}
\includegraphics[width=1.0\linewidth]{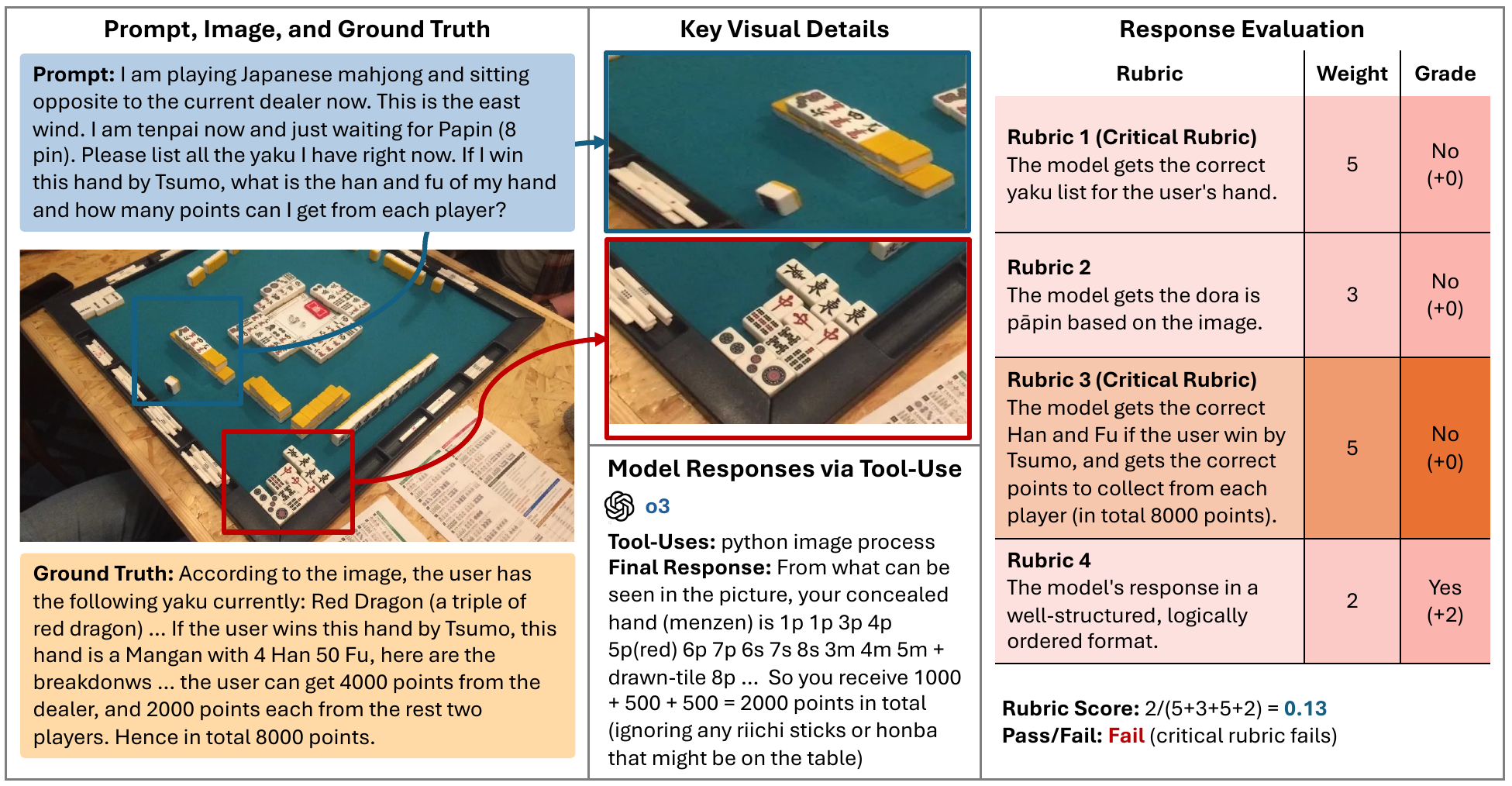}
\end{center}
\caption{Demonstration example from \textsc{VisualToolBench} (single-turn, generalist domain, region switch Q\&A).}
\label{fig:demo_example2}
\end{figure}

\begin{figure}[h]
\begin{center}
\includegraphics[width=1.0\linewidth]{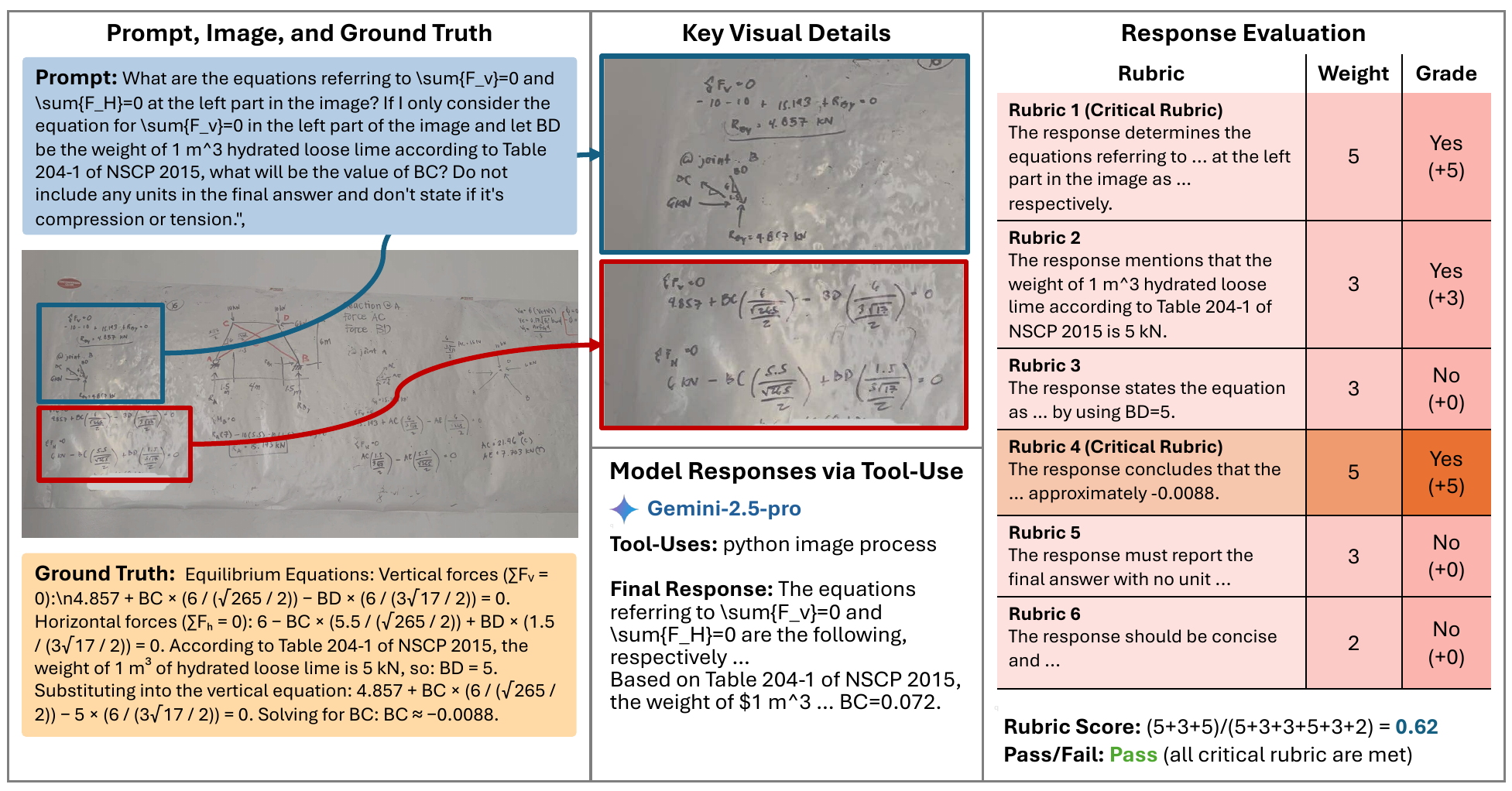}
\end{center}
\caption{Demonstration example from \textsc{VisualToolBench} (single-turn, STEM domain, region switch Q\&A).}
\label{fig:demo_example3}
\end{figure}

In this section, we provide more demonstration examples of \textsc{VisualToolBench}. Figure \ref{fig:demo_example2} and Figure  \ref{fig:demo_example3} provide two more examples with model responses and grading process. 

Below, we present additional benchmark examples covering both single-turn and multi-turn settings across all five task categories. Each example includes the task prompt, image input, golden answer, reference tool trajectory, and evaluation rubrics.

\begin{tcolorbox}[
  enhanced,
  breakable,
  colback=gray!70!gray!3!white,
  colframe=gray!70!gray!60!white,
  title={Single-Turn, Generalist, Hybrid-Reasoning},
  boxrule=0.5mm,
  width=\textwidth,
  arc=1mm,
  auto outer arc,
  before skip=6pt, after skip=6pt
]

\textbf{Prompt:} Check out the menu from this new coffee shop I'm meeting a date at. I want to have a hot coffee ready for him when he gets here. Give me a suggestion for someone who doesn't like steamed milk or sweet-tasting drinks. I don't want it to be the cheapest type of coffee on the menu. 

\vspace{5pt}

\begin{center}
    \includegraphics[width=0.65\textwidth]{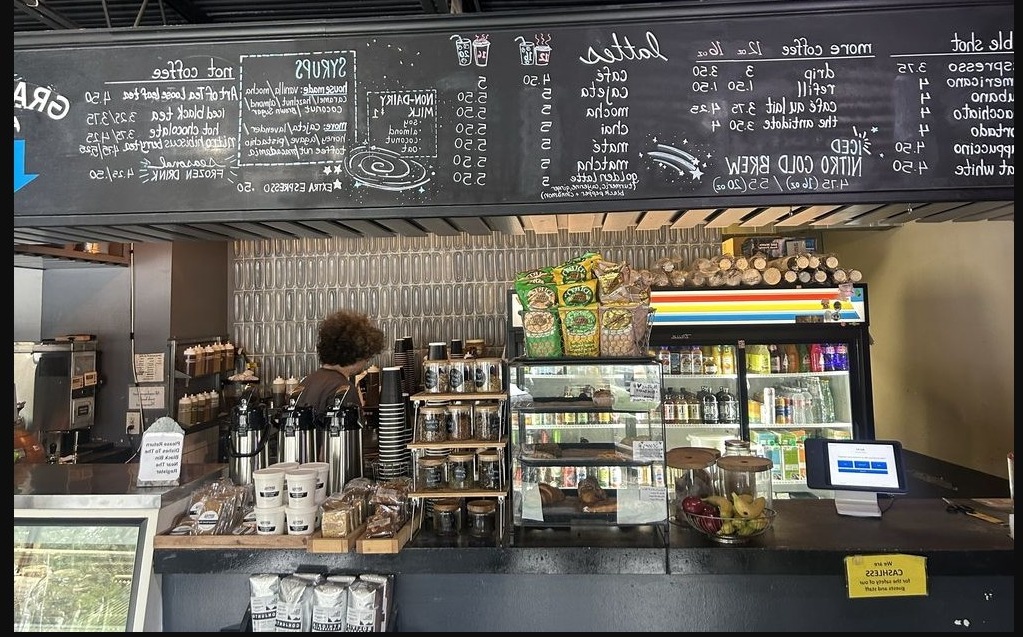}
\end{center}

\textbf{Golden Answer:} That's a great way to start your date! Looking at the menu, an Americano would meet your requirements. It doesn't contain steamed milk, and it's not sweet. At \$4, it's not the cheapest item on the menu, either. Let me know if you need any other suggestions!
\vspace{5pt}

\textbf{Reference Tool Trajectory:}
\begin{enumerate}
    \item \texttt{python\_image\_processing}: Load image to enhance it by flipping horizontally.
    \item \texttt{python\_image\_processing}: Crop the image for items that qualify as coffee and are not the cheapest on the menu.
    \item \texttt{web\_search}: Perform a web search to identify which types of coffee contain steamed milk and are sweet-tasting.
\end{enumerate}
\vspace{5pt}

\textbf{Rubrics:} 

\vspace{4pt}

\small{
\begin{tabular}{p{0.3\linewidth}|c|c|p{0.2\linewidth}|c}
    \toprule
    \textbf{Description} & \textbf{Weight} & \textbf{Type} & \textbf{Category} & \textbf{Critical} \\
    \midrule
    Suggests Americano coffee as the beverage. 
    & 5 & Objective & Reasoning, Instruction following, Truthfulness, Visual Understanding & Yes \\
    \midrule
    States that the Americano is not sweet and does not contain milk.
    & 5 & Objective & Instruction following, Truthfulness & Yes \\
    \midrule
    States that the Americano is not the cheapest coffee on the menu
    & 4 & Objective & Visual Understanding & No \\
    \bottomrule
\end{tabular}
}

\end{tcolorbox}

\begin{tcolorbox}[
  enhanced,
  breakable,
  colback=gray!70!gray!3!white,
  colframe=gray!70!gray!60!white,
  title={Single-Turn, STEM, Region Switch Q\&A},
  boxrule=0.5mm,
  width=\textwidth,
  arc=1mm,
  auto outer arc,
  before skip=6pt, after skip=6pt
]

\textbf{Prompt:} For the top-left flowchart, what operation updates 'S' inside the loop, and what condition terminates it? Also, for the top-right flowchart, what are the two possible calculations for 'y', and what condition decides which one is performed?

\begin{center}
    \includegraphics[width=0.45\textwidth]{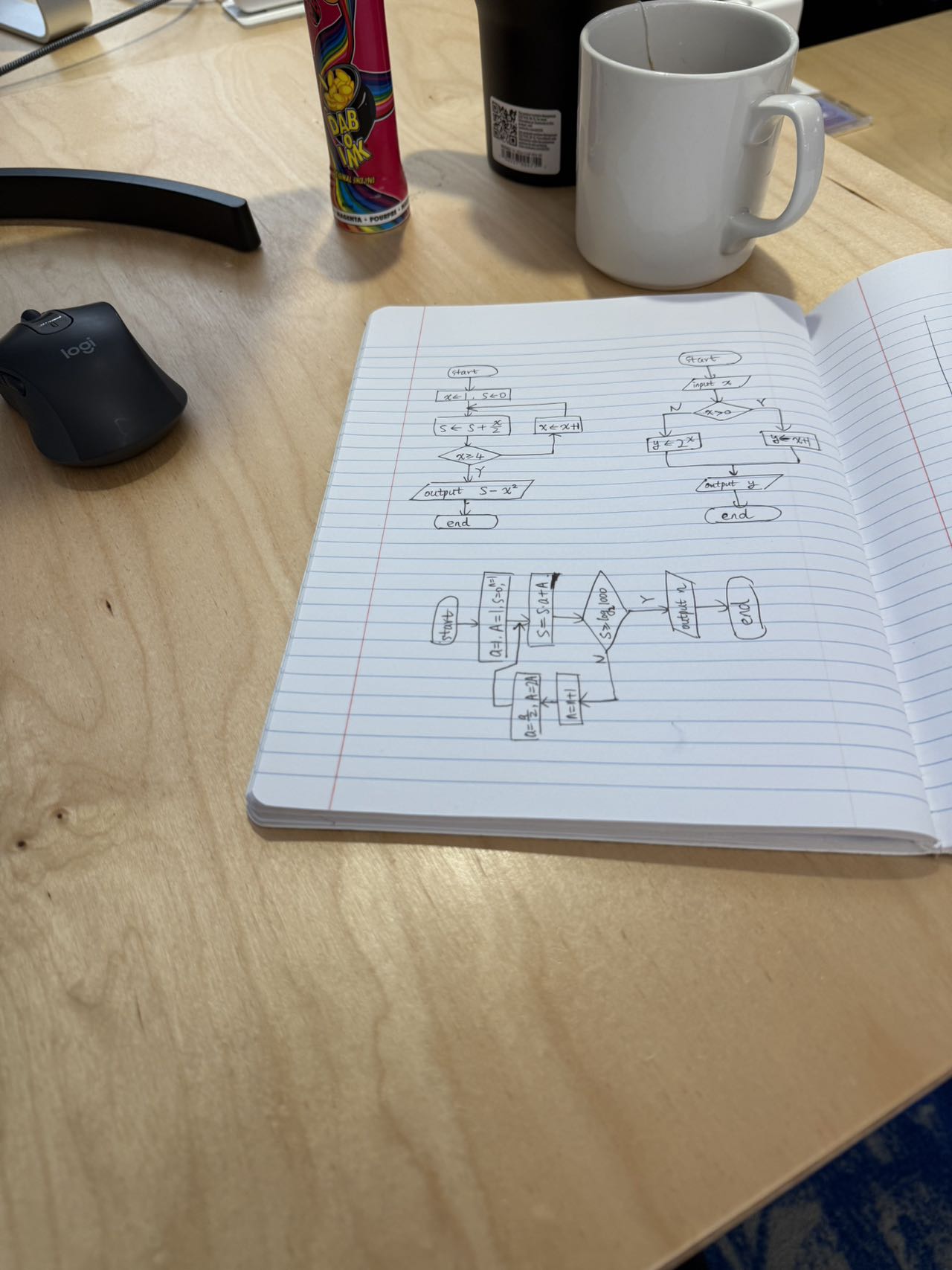}
\end{center}

\textbf{Golden Answer:} The flowchart on the top left defines a loop. The process inside this loop updates the variable 'S' with the specific mathematical operation $S \leftarrow S + x/2$. The loop is controlled by the diamond shape, which sets the exact termination condition to $x \ge 4$. The flowchart on the top-right describes a conditional process based on the input 'x.'. The diamond shape, $x > 0$, is the condition that determines which path is taken. If the condition is true, the calculation performed is $y \leftarrow x + 1$; if it is false, the alternate calculation performed is $y \leftarrow 2^x$.
\vspace{5pt}

\textbf{Reference Tool Trajectory:}
\begin{enumerate}
    \item \texttt{python\_image\_processing}: Crop the top-left flowchart and read the text inside the loop to identify the iterative operation on the 'S' variable ($S \leftarrow S + x/2$) and the text in the diamond shape to identify the loop's termination condition ($x \ge 4$).
    \item \texttt{python\_image\_processing}: Crop the the top-right flowchart and read the text in the diamond to find the deciding condition ($x > 0$) and trace its "Yes" and "No" paths to find the two possible calculations for the output 'y' ($y \leftarrow x + 1$ and $y \leftarrow 2^x$).
\end{enumerate}
\vspace{5pt}

\textbf{Rubrics:}

\vspace{4pt}

\small{
\begin{tabular}{p{0.3\linewidth}|c|c|p{0.2\linewidth}|c}
    \toprule
    \textbf{Description} & \textbf{Weight} & \textbf{Type} & \textbf{Category} & \textbf{Critical} \\
    \midrule
    States that the operation updating $S$ inside the loop is $S \leftarrow S + x/2.$
    & 5 & Objective & Reasoning, Instruction following, Truthfulness, Visual Understanding & Yes \\
    \midrule
    States that the loop terminates when $x > 4$. 
    & 5 & Objective & Reasoning, Instruction following, Truthfulness, Visual Understanding & Yes \\
    \midrule
    The response states both possible calculations for the output $y$ as $y \leftarrow x + 1$ and $y \leftarrow 2^x$. 
    & 4 & Objective & Reasoning, Instruction following, Truthfulness, Visual Understanding & Yes \\
    \midrule 
    The response states that the deciding condition from the top-right flowchart is $x > 0$.
    & 3 & Objective & Reasoning, Instruction following, Truthfulness, Visual Understanding & No \\
    \bottomrule
\end{tabular}
}

\end{tcolorbox}

\begin{tcolorbox}[
  enhanced,
  breakable,
  colback=gray!70!gray!3!white,
  colframe=gray!70!gray!60!white,
  title={Single-Turn, Generalist, Region Switch Q\&A},
  boxrule=0.5mm,
  width=\textwidth,
  arc=1mm,
  auto outer arc,
  before skip=6pt, after skip=6pt
]

\textbf{Prompt:} I need to get to Crossroads Mall entrance from the Central Transfer station by 8:30. When do I need to be ready to be picked up? Also, which road will the route follow for most of the trip?

\begin{center}
    \includegraphics[width=0.6\textwidth]{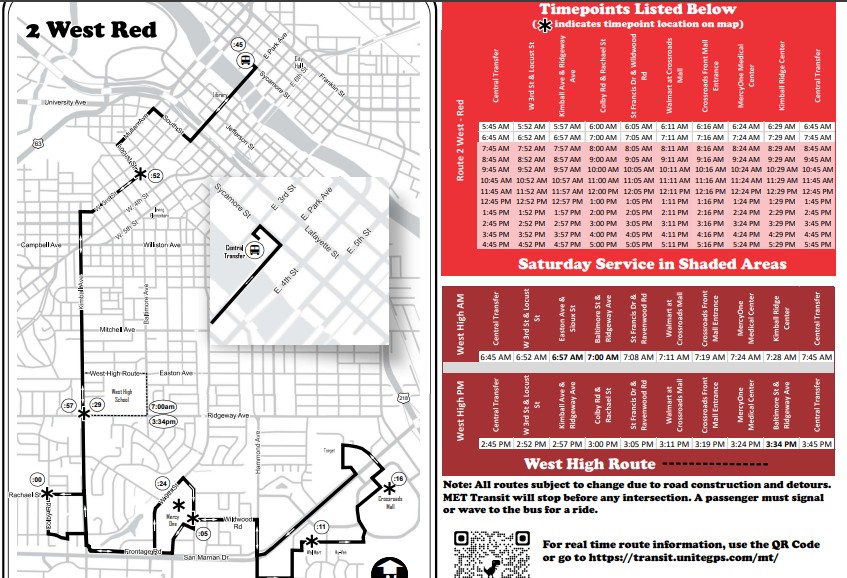}
\end{center}

\textbf{Golden Answer:} 
\begin{itemize}
    \item You will need to be picked up at the Central Transfer station at 7:45 AM, since this bus arrive at Crossroads Mall by 8:16.
    \item The route mainly follows Kimball Avenue from the image.
\end{itemize}
\vspace{5pt}

\textbf{Reference Tool Trajectory:}
\begin{enumerate}
    \item \texttt{python\_image\_processing}: Zoom-in the bus schedule area, find a bus leaving the Central Transfer station to arrive at Crossroads Mall by 8:30. Also use this tool to identify the main road the route follows..
    \item \texttt{python\_image\_processing}: Crop the map area to find the main road name for the bus route.
\end{enumerate}
\vspace{5pt}

\textbf{Rubrics:}

\vspace{4pt}

\small{
\begin{tabular}{p{0.35\linewidth}|c|c|p{0.16\linewidth}|c}
    \toprule
    \textbf{Description} & \textbf{Weight} & \textbf{Type} & \textbf{Category} & \textbf{Critical} \\
    \midrule
    The model identifies the main road the route follows as Kimball Avenue.
    & 5 & Objective & Reasoning, Instruction following, Truthfulness, Visual Understanding & Yes \\
    \midrule
    The response states that the user should be picked up at 7:45 AM to catch the bus arriving at Crossroads Mall entrance by 8:30 AM.
    & 5 & Objective & Reasoning, Instruction following, Truthfulness, Visual Understanding & Yes \\
    \bottomrule
\end{tabular}
}

\end{tcolorbox}

\begin{tcolorbox}[
  enhanced,
  breakable,
  colback=gray!70!gray!3!white,
  colframe=gray!70!gray!60!white,
  title={Multi-Turn, Medical, Temporal Sequential Reasoning},
  boxrule=0.5mm,
  width=\textwidth,
  arc=1mm,
  auto outer arc,
  before skip=6pt, after skip=6pt
]

\begin{center}
\textbf{Turn 1} 
\end{center}

\textbf{Prompt:} The medical team is treating a young patient admitted to the emergency room with a diagnosis of pneumothorax after a MVA. During surgery, a chest drain is placed under fluoroscopic guidance. Based on this image, describe the correct trajectory and placement of the instruments. Also, describe the criteria for confirming the safety of the procedure, as well as the possible complications of improper drain placement.

\begin{center}
    \includegraphics[width=0.5\textwidth]{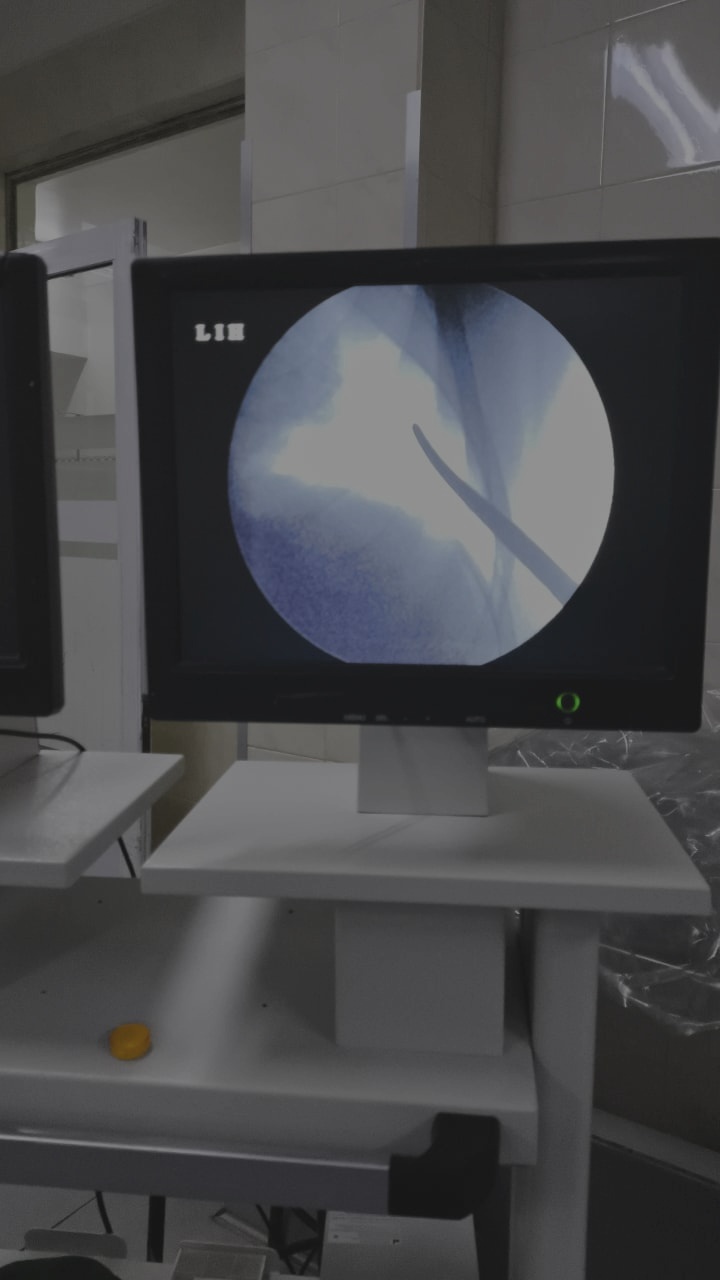}
\end{center}

\textbf{Golden Answer:} 
The image shows the introduction of scissors into the thoracic cavity of a patient with pneumothorax. The instrument's path is carefully directed toward the pleural area, taking great care to avoid direct contact with central structures such as the heart.
The orientation is appropriate, as it follows a lateral and slightly superior path, allowing safe access to the pleural space.
Correct placement and safety criteria: 

\begin{itemize}
\item The instrument is directed toward the pleural cavity and not the mediastinum.
\item It is kept away from the cardiac silhouette and large vessels.
\item The progressive opening of the tract facilitates subsequent placement of the drainage tube.
\end{itemize}

Possible complications of incorrect placement:
\begin{enumerate}
    \item Cardiac or large vessel injury (cardiac tamponade, massive hemothorax).
    \item Lung injury with subcutaneous emphysema or hemothorax.
    \item Ineffective extrapleural placement.
\end{enumerate}

In conclusion, the image shows a technically correct procedure in the scissors dissection phase, maintaining safety by avoiding vital structures before definitive placement of the chest drain.

\vspace{5pt}

\textbf{Reference Tool Trajectory:}

\begin{enumerate}
    \item \texttt{python\_image\_processing}: Apply a brightness and contrast adjustment to clearly display the image. 
    \item \texttt{python\_image\_processing}: Apply a crop operation to extract only the area of the monitor containing the image. 
    \item \texttt{web\_image\_processing}: Search for information about the pneumothorax intervention procedure and its potential complications from improper drain placement.
\end{enumerate}
\vspace{5pt}

\textbf{Rubrics:}
          
\vspace{4pt}

\small{
\begin{tabular}{p{0.35\linewidth}|c|c|p{0.16\linewidth}|c}
    \toprule
    \textbf{Description} & \textbf{Weight} & \textbf{Type} & \textbf{Category} & \textbf{Critical} \\
    \midrule
    Identify that the chest drain orientation follows a lateral and slightly superior path to safely access the pleural space.
    & 5 & Objective & Instruction following, Truthfulness, Visual Understanding & Yes \\
    \midrule
    Identify safety criteria: instrument trajectory into the pleural cavity while avoiding the cardiac silhouette and large vessels. 
    & 5 & Objective & Reasoning, Instruction following, Truthfulness & Yes \\
    \midrule
    Identify possible complications of improper drain placement, including cardiac or vessel injury, lung injury, and ineffective extrapleural placement.
    & 4 & Objective & Reasoning, Instruction following, Truthfulness & Yes \\
    \bottomrule
\end{tabular}
}

\vspace{10pt}

\begin{center}
========================================================================
\end{center}

\begin{center}
\textbf{Turn 2} 
\end{center}

\textbf{Prompt:} The patient also has a fracture due to the MVA, which was treated along with the pneumothorax. The patient returns for routine follow-up after surgical fixation of a fracture. The radiograph shows the surgical site with the internal fixation hardware in place. Identify the position of the fracture and the hardware holding it. Also, identify the progress of bone healing and if there are any potential complications, such as infection, loosening, hardware migration, or malunion.

\begin{center}
    \includegraphics[width=0.5\textwidth]{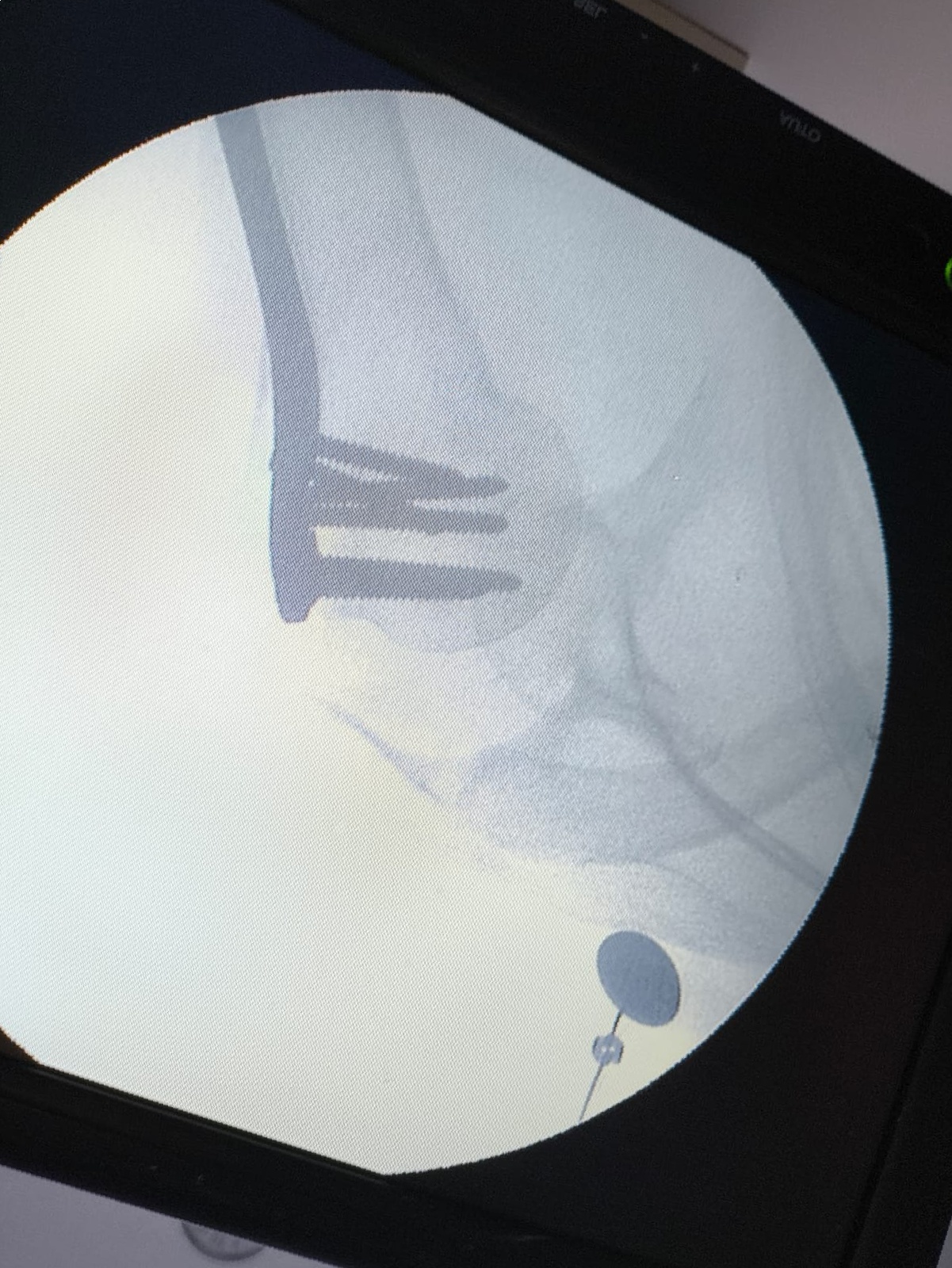}
\end{center}

\textbf{Golden Answer:} 
The follow-up radiograph of the proximal humerus shows internal fixation with a plate and multiple screws appropriately placed in the fracture. The hardware is intact, with no signs of loosening, migration, or fracture. The fracture is still slightly appreciable but shows early union with callus formation, consistent with a normal healing process at this stage. There is no evidence of infection, osteolysis, or malunion. Overall, the postoperative outcome of the proximal humerus fracture is favourable, with correct alignment and satisfactory progression.
\vspace{5pt}

\textbf{Reference Tool Trajectory:}
\begin{enumerate}
    \item \texttt{python\_image\_processing}: Apply a flip operation to flip it vertically and extract it in the correct orientation. 
    \item \texttt{python\_image\_processing}: Apply a rotation operation using the estimated coordinates to extract only the laptop screen area with the imaging findings. 
\end{enumerate}
\vspace{5pt}

\textbf{Rubrics:}

\vspace{4pt}

\small{
\begin{tabular}{p{0.35\linewidth}|c|c|p{0.16\linewidth}|c}
    \toprule
    \textbf{Description} & \textbf{Weight} & \textbf{Type} & \textbf{Category} & \textbf{Critical} \\
    \midrule
    Identifies absence of infection and other complications.
    & 5 & Objective & reasoning, truthfulness, visual understanding & Yes
    \\
    \midrule
    Identify that the proximal humerus fracture is treated with internal fixation hardware (a plate and multiple screws) appropriately placed.
    & 5 & Objective & visual understanding & Yes \\
    \midrule
    Identifies that callus is forming at the fracture site. 
    & 4 & Objective & Reasoning, Visual Understanding & Yes \\
    \midrule
    Identify that this is a shoulder radiograph.
    & 2 & Objective & Visual Understanding & No \\
    \midrule
    Identifies that the radiograph image quality is sufficient for evaluation 
    & 1 & Objective & Visual Understanding & No \\
    \bottomrule
\end{tabular}
}

\end{tcolorbox}

\begin{tcolorbox}[
  enhanced,
  breakable,
  colback=gray!70!gray!3!white,
  colframe=gray!70!gray!60!white,
  title={Multi-Turn, Generalist, Follow-up Test},
  boxrule=0.5mm,
  width=\textwidth,
  arc=1mm,
  auto outer arc,
  before skip=6pt, after skip=6pt
]

\begin{center}
\textbf{Turn 1} 
\end{center}

\textbf{Prompt:} Shown is a map of multiple train lines throughout a city, with each station/stop marked by a white dot (with intersecting lines indicating where one can transfer lines). Suppose I start at the northmost station of Line Seven; then, without using external sources, what is the least number of required stops to reach the southmost station of Line One? Give all the reasoning necessary to solve the question.

\begin{center}
    \includegraphics[width=0.6\textwidth]{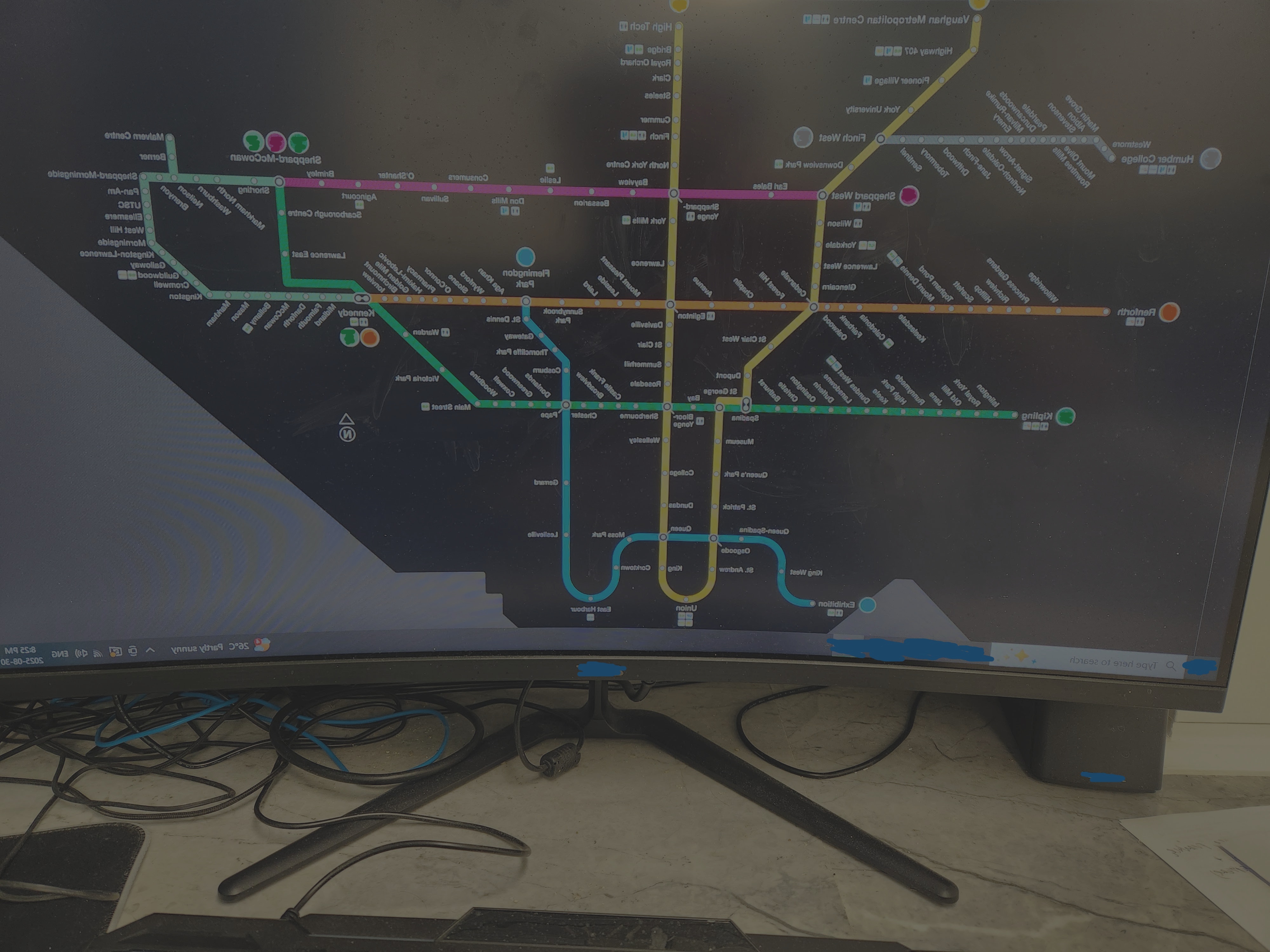}
\end{center}

\textbf{Golden Answer:} 
I am sorry, but it appears the numbers corresponding to each train line are not listed in the image given. To determine both the northmost station of Line Seven, as well as the southmost station of Line One, the line number corresponding to each color (or at least the colors associated with Lines One and Seven) must be given. Can you please provide this information?

\vspace{5pt}

\textbf{Reference Tool Trajectory:}

\begin{enumerate}
    \item \texttt{python\_image\_processing}: Flip the image horizontally, and increase brightness and contrast as needed.
    \item \texttt{python\_image\_processing}: Locate the train line map on the screen of the computer in the upper/middle portion of the image. Identify that the line numbers are not listed/are missing from the picture. 
\end{enumerate}
\vspace{5pt}

\textbf{Rubrics:}
          
\vspace{4pt}

\small{
\begin{tabular}{p{0.35\linewidth}|c|c|p{0.16\linewidth}|c}
    \toprule
    \textbf{Description} & \textbf{Weight} & \textbf{Type} & \textbf{Category} & \textbf{Critical} \\
    \midrule
    The model identifies that the line numbers corresponding to each color are not provided in the image and are necessary to solve the problem.
    & 5 & Objective & Visual Understanding & Yes \\
    \midrule
    The model asks a follow-up clarification question requesting the mapping between line numbers and their colors for Lines One and Seven.
    & 4 & Objective & Instruction following & Yes \\
    \bottomrule
\end{tabular}
}

\vspace{10pt}

\begin{center}
========================================================================
\end{center}

\begin{center}
\textbf{Turn 2} 
\end{center}

\textbf{Prompt:} That's my bad, I forgot to provide you with the line numbers on the map. Can you now please solve the same question? Shown is a map of multiple train lines throughout a city, with each station/stop marked by a white dot (with intersecting lines indicating where one can transfer lines). Suppose I start at the northmost station of Line Seven; then, without using external sources, what is the least number of required stops to reach the southmost station of Line One? Give all the reasoning necessary to solve the question.
\begin{center}
    \includegraphics[width=0.6\textwidth]{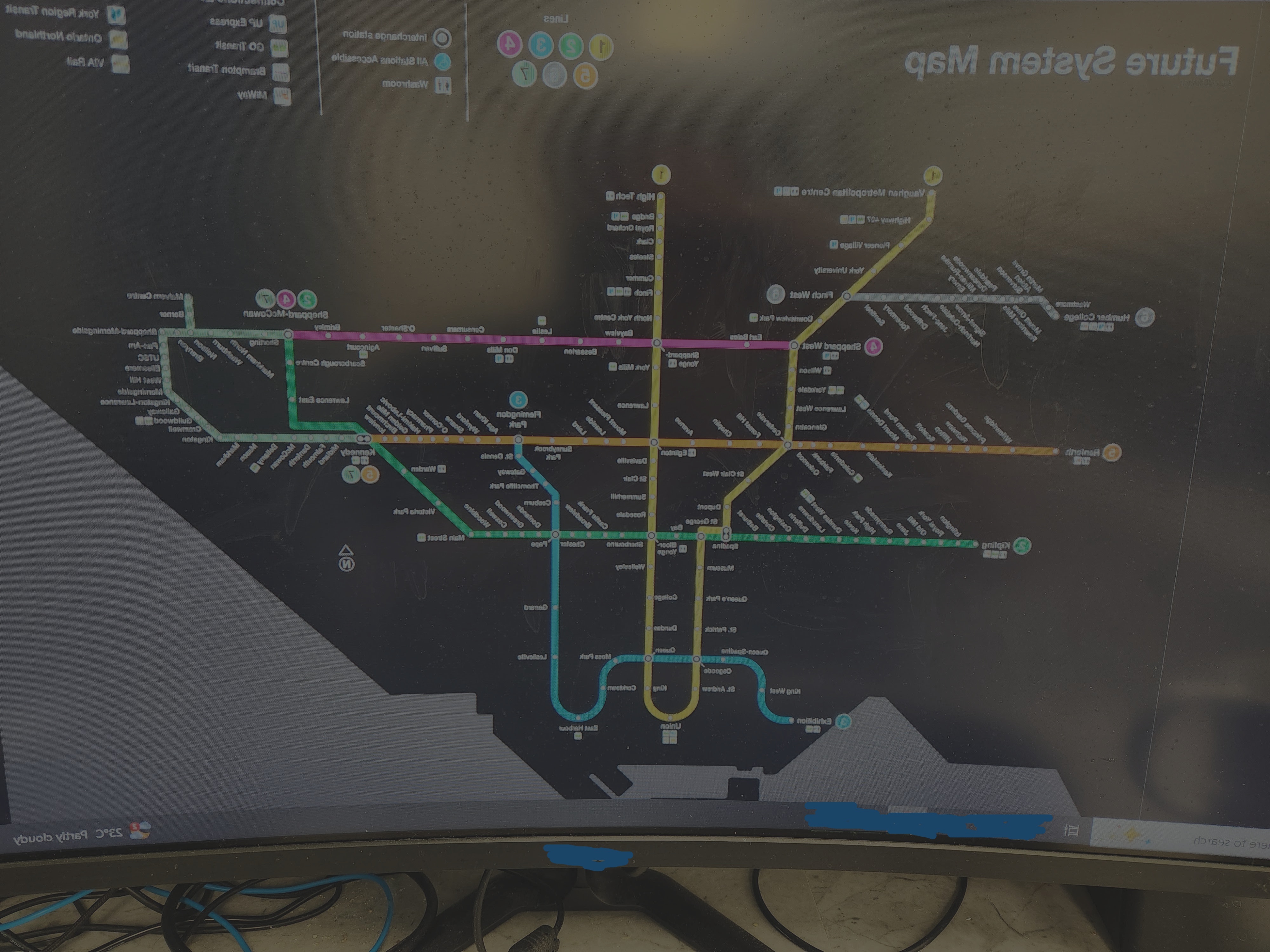}
\end{center}

\textbf{Golden Answer:} 
Using the compass at bottom right (north is up), identify Line 1 (yellow) and Line 7 (light green). The northernmost Line 7 station is Malvern Centre and the southernmost Line 1 station is Union. The minimum-stop route is: Malvern Centre $\to$ Neilson $\to$ west via Washburn $\to$ Sheppard McCowan $\to$ Kennedy $\to$ south to Warden $\to$ Pape $\to$ Garrard $\to$ Queen $\to$ Union, avoiding longer branches via Brenyon or Ionview. Counting stops on this path (excluding the starting station) gives 25.

\vspace{5pt}

\textbf{Reference Tool Trajectory:}

\begin{enumerate}
    \item \texttt{python\_image\_processing}: Flip the image horizontally, and increase brightness and contrast as needed.
    \item \texttt{python\_image\_processing}: Locate the train line map on the computer screen in the middle/upper portion of the image. Further, zoom into the upper portion of the screen to identify the train line color pairing as: light green (Line Seven) and yellow (Line One). Next, locate the compass symbol near the bottom right portion of the screen, indicating that the top of the image is the northmost portion. Finally, identify all possible station stops (marked as white dots) between the northmost station of Line Seven (Malvern Centre) and the southmost station of Line One (Union).
\end{enumerate}
\vspace{5pt}

\textbf{Rubrics:}

\vspace{4pt}

\small{
\begin{tabular}{p{0.35\linewidth}|c|c|p{0.16\linewidth}|c}
    \toprule
    \textbf{Description} & \textbf{Weight} & \textbf{Type} & \textbf{Category} & \textbf{Critical} \\
    \midrule
    Model correctly identifies the northmost station of Line Seven as Malvern Centre.
    & 5 & Objective & Reasoning, Truthfulness, Visual Understanding & Yes \\
    \midrule
    Model correctly identifies the southmost station of Line One as Union Station.
    & 5 & Objective & Reasoning, Instruction following, Truthfulness, Visual Understanding & Yes \\
    \midrule
    The model states that the minimum number of stops required when travelling from Malvern Centre to Union Station is 25.
    & 4 & Objective & Reasoning, Instruction following, Truthfulness, Visual Understanding & Yes \\
    \midrule
    The model outlines a path listing stations from Malvern Centre to Union Station that follows the specified route, allowing equivalent station names.
    & 3 & Objective & Reasoning, Instruction following, Truthfulness, Visual Understanding & No \\
    \bottomrule
\end{tabular}
}

\end{tcolorbox}

\begin{tcolorbox}[
  enhanced,
  breakable,
  colback=gray!70!gray!3!white,
  colframe=gray!70!gray!60!white,
  title={Multi-Turn, Sports, Progressive Visual Reasoning},
  boxrule=0.5mm,
  width=\textwidth,
  arc=1mm,
  auto outer arc,
  before skip=6pt, after skip=6pt
]

\begin{center}
\textbf{Turn 1} 
\end{center}

\textbf{Prompt:} A badminton tournament during the 2016 Rio Olympics between Lin Dan (in red) and Lee Chong Wei (in blue) is shown, which is a best-of-three-game series. The middle of a round is currently in play. Who is currently hitting/about to hit the shuttlecock, and how do you know? Further, if the shuttlecock is missed by the opponent after being hit by the player from the answer in the first question, what numbered point on the ground labelled in red will be the farthest point away from the receiver (player who missed), such that the shuttlecock is not considered out?

\begin{center}
    \includegraphics[width=0.5\textwidth]{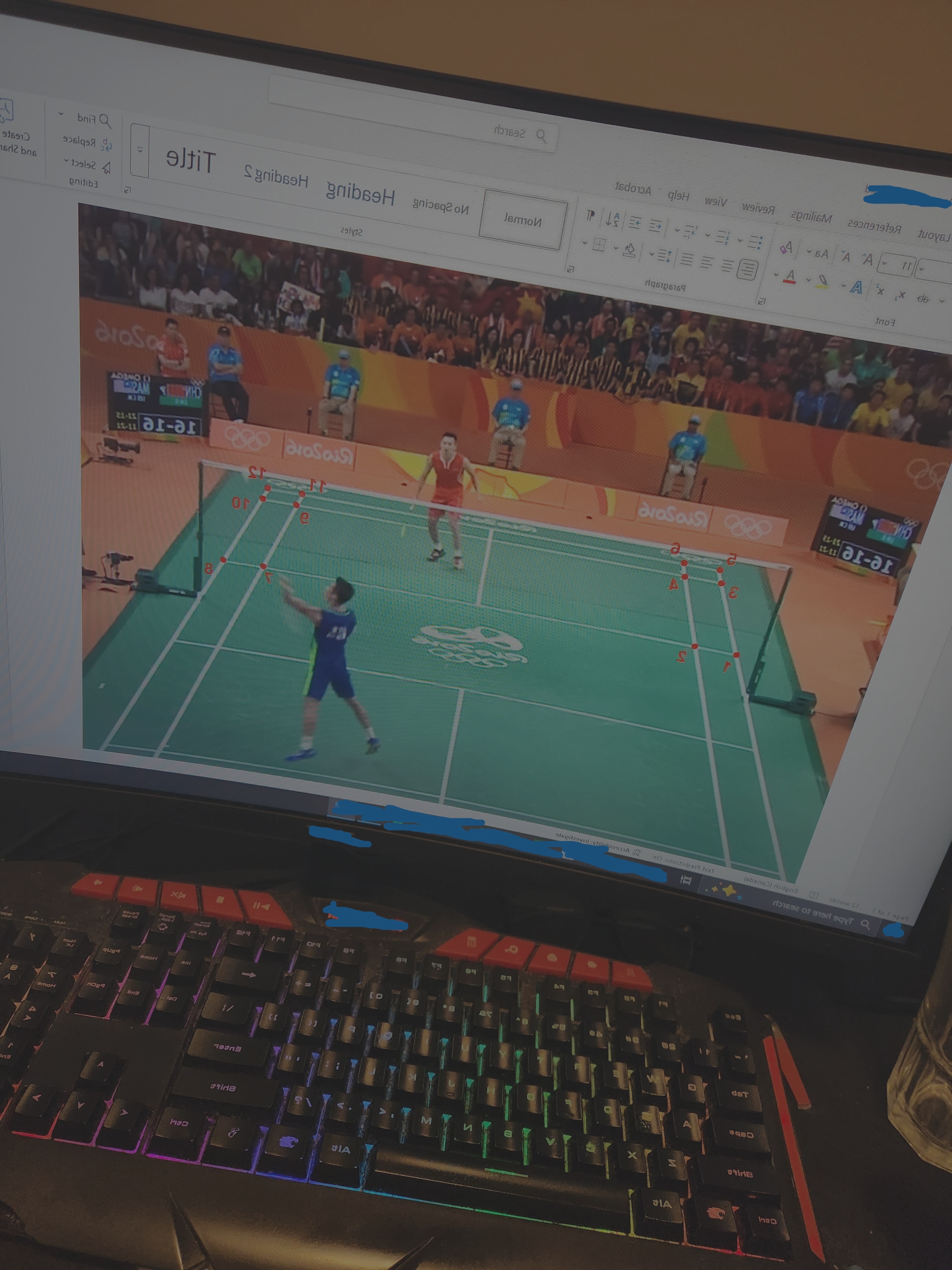}
\end{center}

\textbf{Golden Answer:} 
\textbf{Who is hitting/about to hit the shuttlecock?}
Since the player in blue is jumping, this is a clear indication that he is currently hitting/about to hit the shuttlecock. Further, if one looks more carefully, a white shuttlecock can be identified near the top right of the blue player's racket. Thus, the blue player (Lee Chong Wei) is currently hitting/about to hit the shuttlecock.

\textbf{What point labelled in red is farthest away from the receiving player if he misses the shot made by Lee Chong Wei, not considered out?}
Since the game shown is between two players (singles), any shots where the shuttlecock lands in the farthest left or farthest right rectangles are considered out (everything else is considered in, including the backmost rectangle). Thus, this eliminates the points numbered 1, 3, 5, 8, 10, and 12 as possible answers. Of the remaining points, the point numbered 6 is furthest away from Lin Dan (red player). Thus, the point labelled 6 is the farthest point away from the receiving player (Lin Dan) considered in, assuming he misses the shot made by Lee Chong Wei.

\vspace{5pt}

\textbf{Reference Tool Trajectory:}

\begin{enumerate}
    \item \texttt{python\_image\_processing}: Flip the image horizontally.
    \item \texttt{python\_image\_processing}: Adjust the brightness and contrast of the image.
\end{enumerate}
\vspace{5pt}

\textbf{Rubrics:}
          
\vspace{4pt}

\small{
\begin{tabular}{p{0.35\linewidth}|c|c|p{0.16\linewidth}|c}
    \toprule
    \textbf{Description} & \textbf{Weight} & \textbf{Type} & \textbf{Category} & \textbf{Critical} \\
    \midrule
    The model correctly identifies Lee Chong Wei (in blue) as the player currently hitting or about to hit the shuttlecock.
    & 5 & Objective & Instruction following, Truthfulness, Visual Understanding, Reasoning & Yes \\
    \midrule
    The model justifies its identification by referencing either the player's posture/position or the visible shuttlecock location near the racket.
    & 4 & Objective & Visual Understanding, Reasoning & Yes \\
    \midrule
    The model must reason that the game shown is a singles match between two players.
    & 3 & Objective & Reasoning & No \\
    \midrule
    The model must state that the farthest most left and farthest most right rectangles are considered out if the shuttlecock lands there.
    & 3 & Objective & Reasoning & No \\  
    \midrule
    Identify the 12 points labelled in red (1 through 12) on Lin Dan's side of the court.
    & 1 & Objective & instruction following, Visual Understanding & No \\  
    \midrule
    State that the shuttlecock is out if it lands in any of the labelled points 1, 3, 5, 8, 10, or 12.
    & 2 & Objective & instruction following, reasoning, truthfulness & No \\  
    \midrule
    State that point number 6 is the farthest allowed point from Lin Dan still considered in..
    & 5 & Objective & Instruction following, Truthfulness, Visual Understanding, Reasoning & Yes \\  
    \midrule
    Response includes explicit reasoning rather than only providing a ground truth answer.
    & 3 & Objective & Reasoning, Presentation & No \\  
    \bottomrule
\end{tabular}
}

\vspace{10pt}

\begin{center}
========================================================================
\end{center}

\begin{center}
\textbf{Turn 2} 
\end{center}

\textbf{Prompt:} Suppose now, instead of hitting the shuttlecock to the point labelled 6 on the ground, Lee Chong Wei instead hits the shuttlecock to the point labelled 8 on the ground. Assuming Lin Dan does not receive (hit) the shuttlecock and lets it hit the ground, what will the scoreboard read after this point? State the player's name associated with the number of points as well.
\vspace{10pt}

\textbf{Golden Answer:} 
\textbf{Current score according to the scoreboard:} From the scoreboard, it reads 16-16, where the left number is the score associated with the team from China (from the flag shown), whereas the right number is the score associated with the team from Malaysia (from the flag shown).

\textbf{New score, assuming the shuttlecock lands at the point labelled 8:} Since the game is a singles match, the outer left and outer right-most rectangles are considered out. Since the point labelled 8 is in the outer right-most rectangle, this shot made by Lee Chong Wei would be considered out. Since Lin Dan did not receive the shot as stated by the prompt, Lin Dan will win this round, increasing the score to 17-16. (Here, it is important to note that the left number on the scoreboard has increased, and not the right). Further, Lin Dan will have 17 points, while Lee Chong Wei will still have 16 points.

\vspace{5pt}

\textbf{Reference Tool Trajectory:}

\begin{enumerate}
    \item \texttt{python\_image\_processing}: Identify the current match score as 16-16 (middle top right of the image), where the left number is associated with the score of the team from China, whereas the right number is associated with the score of the team from Malaysia, according to the flags.
    \item \texttt{web\_search}: Look up and determine that Lin Dan represents the team from China, whereas Lee Chong Wei represents the team from Malaysia.
\end{enumerate}
\vspace{5pt}

\textbf{Rubrics:}

\vspace{4pt}

\small{
\begin{tabular}{p{0.35\linewidth}|c|c|p{0.16\linewidth}|c}
    \toprule
    \textbf{Description} & \textbf{Weight} & \textbf{Type} & \textbf{Category} & \textbf{Critical} \\
    \midrule
    Identify the current score as 16-16.
    & 3 & Objective &Truthfulness, Visual Understanding & Yes \\
    \midrule
    Associate the left number with the team from China and the right number with the team from Malaysia based on the flags. 
    & 1 & Objective &Truthfulness, Visual Understanding & No \\
    \midrule
    Model correctly assigns Lin Dan to the left scoreboard number (team from China) and Lee Chong Wei to the right scoreboard number (team from Malaysia).
    & 2 & Objective & Reasoning, Visual Understanding & No \\
    \midrule
    Model correctly reasons that because the shuttlecock landing point labelled 8 is out in singles matches and Lin Dan does not hit it back, Lin Dan receives the point 
    & 5 & Objective & Reasoning, Visual Understanding & Yes \\
    \midrule   
    The model states the new score after Lin Dan receives the point as: 17-16 (17 for Lin Dan, and 16 for Lee Chong Wei).
    & 5 & Objective & Reasoning & Yes \\
    \midrule  
    The model states the new score after Lin Dan receives the point as: 17-16 (17 for Lin Dan, and 16 for Lee Chong Wei). 
    & 5 & Objective &Truthfulness, Instruction Following & Yes \\  
    \midrule
    The reasoning explains why point 8 is considered out.
    & 4 & Objective &Reasoning, Presentation & Yes \\  
    \midrule
    The reasoning explains why the left number on the scoreboard is associated with Lin Dan and not Lee Chong Wei. 
    & 1 & Objective &Reasoning, Presentation & No \\  
    \bottomrule
\end{tabular}
}
\vspace{10pt}

\begin{center}
========================================================================
\end{center}

\begin{center}
\textbf{Turn 3} 
\end{center}

\textbf{Prompt:} Now, since the tournament is a best-of-three-game series, if Lee Chong Wei in fact does hit the shuttlecock out at the point labelled 8 on the ground as discussed, how many additional points will he need to win the match (assuming the score does not tie at 20-20)? If Lee Chong Wei wins this match/game, is he guaranteed to win the series? If so, what were the scores of the previous two matches played against each other?

\vspace{10pt}

\textbf{Golden Answer:} 
\textbf{How many points does Lee Chong Wei need to win the match after shooting the shuttlecock out?} The new score after Lee Chong Wei hits the shuttlecock out is 17-16 (17 for Lin Dan, 16 for Lee Chong Wei). Since a standard badminton game is up to 21 points, and assuming the score does not tie to 20-20 (in which case you can only win if you get two points back-to-back), Lee Chong Wei thus needs 21-16=5 points more to win.

\textbf{Scores of the previous two matches:} One can deduce that the previous two games have already been finished from the scoreboard, where it states that Lin Dan won one game with a final score of 21-15, whereas Lee Chong Wei won the other game with a score of 11-21. Thus, the game being played currently is the final game to determine the winner of the three-game series. Hence, if Lee Chong Wei wins this match, he is guaranteed to win the series.

\vspace{5pt}

\textbf{Reference Tool Trajectory:}

\begin{enumerate}
    \item \texttt{python\_image\_processing}: Identify the scores of the previous two matches/games in the middle top right of the image as: 21-15 (for Lin Dan), and 11-21 (for Lee Chong Wei).
    \item \texttt{web\_search}: Lookup the amount of points needed to win a standard badminton game as 21 (assuming a 20-20 point tie does not occur).
\end{enumerate}
\vspace{5pt}

\textbf{Rubrics:}

\vspace{4pt}

\small{
\begin{tabular}{p{0.35\linewidth}|c|c|p{0.16\linewidth}|c}
    \toprule
    \textbf{Description} & \textbf{Weight} & \textbf{Type} & \textbf{Category} & \textbf{Critical} \\
    \midrule
    Identifies the final scores of the previous two matches as 21-15 (Lin Dan) and 11-21 (Lee Chong Wei).
    & 3 & Objective & Instruction following, Truthfulness, Visual Understanding & No \\
    \midrule
    The model reasons that Lee Chong Wei must reach a total point count of 21 to win the match (assuming a 20-20 tie does not occur).
    & 4 & Objective & Reasoning & Yes \\
    \midrule
    The model correctly calculates that Lee Chong Wei needs 5 additional points to win the current match.
    & 5 & Objective & Reasoning, Instruction following, Truthfulness  & Yes \\
    \midrule
    The model correctly states that if Lee Chong Wei wins this match, he is guaranteed to win the best-of-three series and provides the scores of the previous two matches.
    & 5 & Objective & Reasoning, Presentation & Yes \\
    \midrule
    The response provides intermediate reasoning that addresses each question posed by the prompt: (1) how many additional points Lee Chong Wei needs to win the match, and (2) whether he is guaranteed to win the series and what the scores of the previous two matches were.
    & 2 & Objective & Reasoning, Presentation & No \\
    \bottomrule
\end{tabular}
}

\end{tcolorbox}


\end{document}